\pdfoutput=1
\documentclass[11pt]{article}

\usepackage[final]{acl}

\usepackage{times}
\usepackage{latexsym}

\usepackage[T1]{fontenc}
\usepackage[utf8]{inputenc}

\usepackage{microtype}

\usepackage{inconsolata}

\usepackage{graphicx}

\usepackage{algorithm}
\usepackage{algpseudocode}
\usepackage{multirow}
\usepackage{xcolor}
\usepackage{color, colortbl}
\definecolor{LightCyan}{rgb}{0.88,1,1}
\usepackage{tabularx}
\usepackage{paralist}
\usepackage{subcaption}
\usepackage{longtable}
\usepackage{adjustbox}
\usepackage{amssymb}% http://ctan.org/pkg/amssymb
\usepackage{pifont}% http://ctan.org/pkg/pifont
\usepackage{makecell}
\newcommand{\thickhline}{\noalign{\hrule height 1.5pt}}

\usepackage{hyperref}
\usepackage{booktabs}

\title{QG-CoC: Question-Guided Chain-of-Captions for Large Multimodal Models}

\author{Kuei-Chun Kao$^1$, Hsu Tzu Yin$^1$, Yunqi Hong$^1$, Ruochen Wang$^1$, Cho-Jui Hsieh$^1$ \\
  $^1$Department of Computer Science, University of California, Los Angeles \\
  \texttt{johnson0213@g.ucla.edu} \quad
  \texttt{chohsieh@cs.ucla.edu} \\ \\
\href{https://johnsonkao0213.github.io/QG-CoC/}{https://johnsonkao0213.github.io/QG-CoC/}}

\begin{document}
\maketitle

\begin{abstract}
% Chain-of-Thought (CoT) prompting has proven highly effective for enhancing complex reasoning in Large Language Models (LLMs) and Multimodal Large Language Models (MLLMs). 
Recently, Multimodal Large Language Models (MLLMs) encounter two key issues in multi-image contexts: (1) a lack of fine-grained perception across disparate images, and (2) a diminished capability to effectively reason over and synthesize information from multiple visual inputs.
However, while various prompting methods aim to describe visual content, many existing studies focus primarily on single-image settings or specific, constrained scenarios. This leaves a critical gap in understanding and addressing how MLLMs tackle more general and complex multi-image reasoning tasks. 
Thus, we first extensively investigate how current prompting methods perceive fine-grained visual details and process visual information when dealing with multiple images. Our findings reveal that existing prompting methods fall short in attending to needed clues and seamlessly integrating perception and reasoning. Inspired by the findings, we propose a new zero-shot prompting method, Question-Guided Chain-of-Captions (QG-CoC), a generalized prompting approach that effectively handles problems with an arbitrary number of images. We evaluate our method on various open-source and closed-source MLLMs for multi-image and single-image benchmarks. Experimental results indicate that QG-CoC demonstrates competitive performance across tasks and exhibits robust improvements in the challenging scenarios where existing prompting methods fail.

% {\color{red}(cho: I feel we don't emphasize multi-image enough in the story. 

% Also, I'm slightly confused by the current story. Are we saying currently there's no good prompting methods for multi-image tasks so we propose a new prompting method that achieves sota? I remember the previous plan was to say there are prompting methods to describe image but most of those studies are in single image setting, so we propose a comprehensive study in multi-image setting and then propose our method. \color{blue} (Johnson: Yeah, in section 3, we will first do comprehensive study in multi-image setting such as how to effectively caption image to provide as context to answer multi image question and we adopt single image prompting method to multi image scenario, and see the result in section 3 and 4 (as comparison with our method). )}

\end{abstract}
% the research focuses on two aspects: to evaluate whether LMMs can effectively reason and pair relevant images), and second, multi-image-to-text matching (to assess whether LMMs can accurately capture and summarize detailed image information)
\section{Introduction}
Recent advancements in MLLMs~\cite{li2024llava, liu2023visual} have demonstrated impressive abilities in understanding the semantics of multimodal data and achieving promising results across various single-image tasks. However, recent empirical studies~\cite{meng2024mmiu} show that MLLMs currently still struggle with solving complex multimodal understanding tasks such as temporal, spatial, and multi-image relationships. 
% muir

% self-consistency, Formulate and Solve, Plan and Solve, Self-revision
% To enhance the reasoning ability of LLMs in many tasks (e.g., math reasoning, theorem proving, and advanced problem-solving capabilities), CoT prompting~\cite{wei2022chain} has been studied extensively in the text domain. 
Therefore, there have been some emerging prompting methods that help to enhance the reasoning chain of multimodal data. Most of the works focus on converting visual scenes into rich text-based representations such as scene graph, visual table, and bounding box % textcot, visual-cot
detection~\cite{mitra2024ccot, shao2024visualcot}, then triggering the reasoning ability of MLLMs.
Although these methods are effective for understanding single-image context, they encounter obstacles when discerning relationships between multiple images. This difficulty primarily stems from an insufficient focus on key information, which requires joint consideration of all images involved. Although some methods~\cite{zhang2024cocot} start to consider multiple images in their prompting methods, they are far from being general and dealing with different kinds of scenarios that involve multi-perspectives, multi-relations, and multi-understanding~\cite{wang2024muirbench, meng2024mmiu}. 

In our preliminary study, we first conduct a comprehensive evaluation of various captioning strategies to analyze how to caption images effectively under multi-image scenarios. Our findings reveal that question-guided captioning each image in detail benefits more than captioning multiple images as a whole or concisely. Then, we adopt existing prompting methods to multi-image scenarios and observe the limitations of existing methods that generate a lack of spatial context, unrelated object descriptions, and vague descriptions.
Motivated by our preliminary study, we propose \textbf{QG-CoC}, which first decomposes the original question into necessary sub-questions to understand which key information is needed for solving different tasks. Then, based on each specific sub-question, we generate relevant captioning to ensure each caption is conditioned under the given sub-question. After obtaining guided captions, we utilize each sub-caption as a clear hint to answer each sub-problem. Last, we combine the sub-question and sub-answer pairs to serve as prior domain knowledge, highlighting the key information needed to generate a final response.

\begin{figure*}[ht]
\centering
\includegraphics[width=0.88\textwidth]{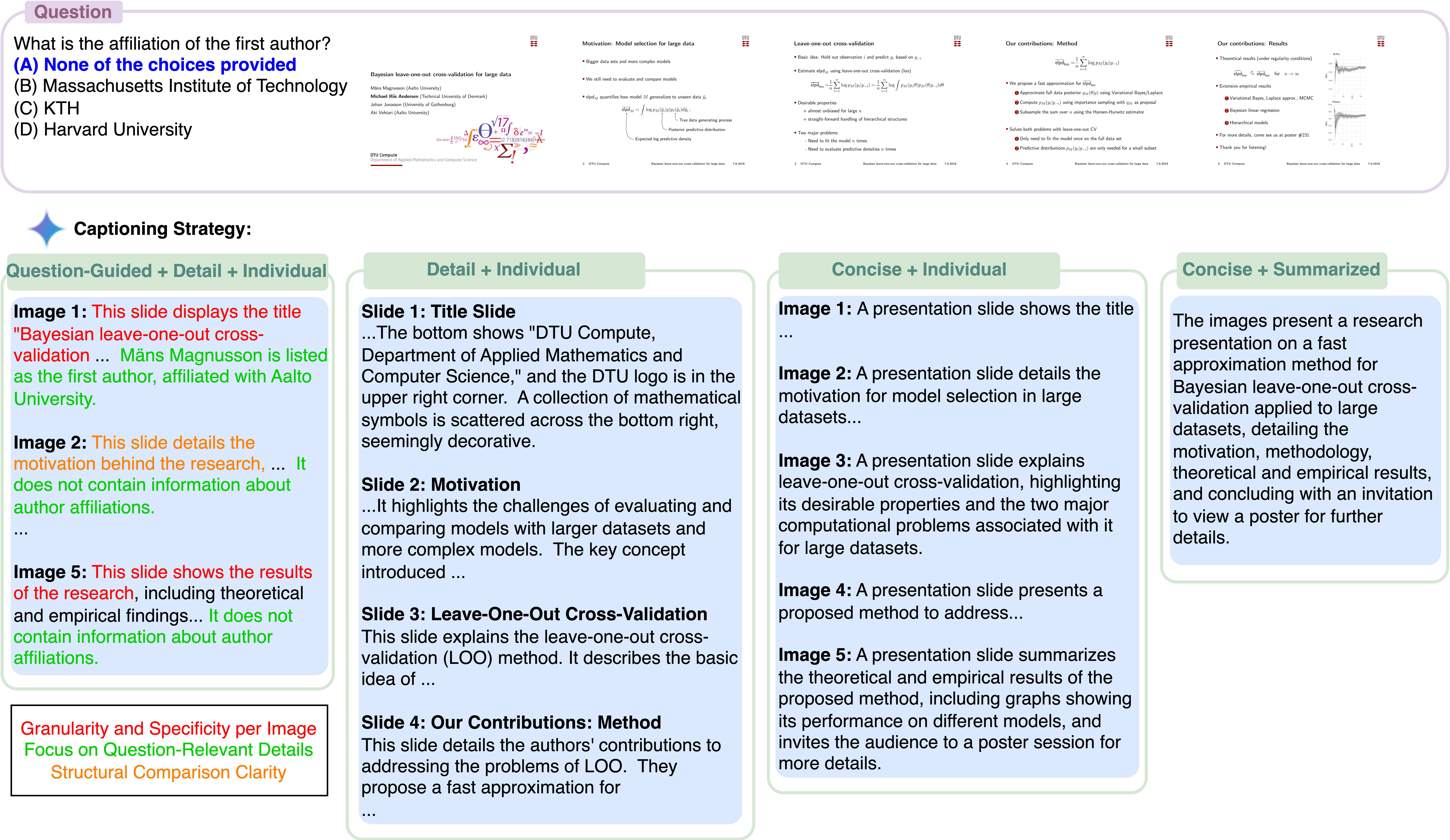}
\vspace{-2mm}
\captionsetup{font=small, width=0.9\linewidth}
\caption{An example multi-image question with different captioning settings. Text in red, green, and orange highlights our advantages. Text in blue is the correct answer. The actual prompt used for each captioning setting can be found in Appendix~\ref{sec:different_captioning_strategy}.}
\label{fig:Caption_Settings}
\vspace{-2mm}
\end{figure*}

To summarize, our main contributions are as follows: 
\begin{compactitem}
    \item We first analyze why existing prompting methods cannot work and suggest what is the most effective way to caption images under multi-image scenarios. 
    \item We then introduce QG-CoC, a novel zero-shot prompting method that can deal with an arbitrary number of images. This provides a strong baseline for future multimodal understanding tasks.
    \item Our method consistently outperforms existing prompting methods in multi-image scenarios and also shows generalization in single-image scenarios under both closed-source and open-source models.
\end{compactitem}

\section{Related Work}
\paragraph{MultiModal Prompting Methods.}
Chain-of-Thought (CoT) prompting has considerably enhanced the reasoning capacities of LLMs. Recent research has explored various methodologies to adapt CoT for multimodal models. Some investigations adopt a two-stage approach, where image information is initially transformed and grounded into captions, graph structure (e.g., scene graphs or knowledge graphs), or bounding boxes before reasoning~\cite{mitra2024ccot, zhang2024cocot, shao2024visualcot, zhang2023multimodal, mondal2024kam, zhong2024visualtable}. 
Other studies use agent-style pipelines that integrate external tools to process and reason with image observations. These tools include code interpreters and specialized vision models~\cite{shao2024visualcot, lei2024scaffolding, hu2024visual, gao2024cantor}.
Although these approaches effectively manage both textual and visual input, they exhibit limitations in handling multi-image scenarios since they need models to automatically integrate and analyze either spatial, temporal, or contextual cues from varied perspectives, moments, and settings~\cite{shao2024visualcot}. To address these limitations, in our work, a general prompting framework is designed for multimodal reasoning without fine-tuning or relying on separate visual modules or external tools.

\paragraph{MultiModal Understanding Benchmarks.}
There are lots of benchmarks have been developed to comprehensively assess the multimodal understanding and reasoning capabilities of MLLMs that require conditioning on images; however, they predominantly focus on single-image scenarios and do not directly measure how well the model and the prompting methods can integrate information across different images~\cite{yue2024mmmu, liu2024mmbench, lu2022scienceqa}. Therefore, several benchmarks have recently been introduced to systematically evaluate multi-image reasoning and understanding capabilities, covering diverse perspectives and tasks such as comparison, video understanding, and grounding~\cite{wang2024muirbench, meng2024mmiu}. Besides, these benchmarks comprehensively assess MLLMs, covering a broader range of current multi-image capacities. Despite these efforts, existing MLLMs fail to explore and unlock the inherent reasoning capabilities without specific prompting to solve multi-image problems, and most of the common techniques to enhance performance are based on supervised fine-tuning~\cite{liu2023visual, jiang2024mantis, xu2024llava} on multi-image interleaved data or CoT reasoning data. In parallel, in our work, we focus on how to apply a sophisticated prompting strategy without fine-tuning to represent visual scenes into more informative descriptions, demonstrating benefits in diverse domains in both single-image and multi-image scenarios.

\section{Preliminaries}
% To effectively address the limitations of Vision-Language Models (VLMs) in handling complex compositional visual-textual relationships, we introduce CECE: Caption Expansion with Contradictions and Entailments. Our approach leverages Natural Language Inference (NLI) to systematically generate entailments and contradictions for each image-caption pair, capturing the deeper meaning of the text and a more interpretable metric for image-to-text and text-to-image evaluation and alignment. We describe our proposed method, beginning with the generation of entailments and contradictions via CECE (Section 3.1). We then explain the likelihood computation for each caption expansion (Section 3.2). Finally, we describe the score-balancing mechanism we use to integrate the contributions of entailments, contradictions, and the original captions into a unified evaluation framework (Section 3.3).

\begin{table*}[ht]
\centering
\resizebox{0.88\textwidth}{!}{%
\begin{tabular}{p{4cm}cccccc}
\toprule
\multicolumn{1}{l}{\textbf{Model}} & \multicolumn{2}{c}{\textbf{Gemini-Flash}} & \multicolumn{2}{c}{\textbf{LLaVA-OV}} & \multicolumn{2}{c}{\textbf{Mantis}} \\ \midrule
Dataset & MMIU & MUIR & MMIU & MUIR & MMIU & MUIR \\
\midrule
Concise vs. Detailed        & 54.1 $\rightarrow$ \textbf{54.9} & 65.2 $\rightarrow$ \textbf{66.3} & 47.3 $\rightarrow$ \textbf{48.0} & 43.7 $\rightarrow$ \textbf{44.0} & 45.3 $\rightarrow$ \textbf{46.4} & 42.3 $\rightarrow$ \textbf{44.5} \\
Summarize vs. Individual   & 54.1 $\rightarrow$ \textbf{54.5} & 66.0 $\rightarrow$ \textbf{66.5} & 46.5 $\rightarrow$ \textbf{48.6} & \textbf{44.1} $\rightarrow$ 43.9 & 45.3 $\rightarrow$ \textbf{46.4} & 43.1 $\rightarrow$ \textbf{43.5} \\
Question-Guided (N/Y)             & 53.3 $\rightarrow$ \textbf{55.3} & 65.4 $\rightarrow$ \textbf{66.2} & 47.4 $\rightarrow$ \textbf{47.8} & 43.1 $\rightarrow$ \textbf{44.7} & 45.5 $\rightarrow$ \textbf{46.0} & 42.4 $\rightarrow$ \textbf{44.1} \\
% Question Decomposition      & 55.3 $\rightarrow$ \textbf{55.5} & 66.2 $\rightarrow$ \textbf{67.6} & 47.8 $\rightarrow$ \textbf{49.9} & 44.7 $\rightarrow$ \textbf{45.9} & 46.0 $\rightarrow$ \textbf{49.0} & 44.1 $\rightarrow$ \textbf{46.8} \\
\bottomrule
\end{tabular}
}
\captionsetup{font=small, width=0.9\linewidth}
\caption{Comparison of captioning settings across models and multi-image datasets. Metrics represent answer accuracy (\%).}
\vspace{-4mm}
\label{tab:captioning-settings}
\end{table*}

\subsection{Analysis on Different Captioning Strategies under Multi-Image}
MLLMs are capable of reasoning directly over both vision and language modalities. These models typically receive an input consisting of images \( I \) and an associated task prompt in text form \( P \) (e.g., a question, caption generation, or scene graph generation). The diverse descriptions generated from these inputs often encapsulate multiple perspectives and provide advantageous informative context that aids in addressing the original problem. However, a critical question arises: \textit{How can we accurately generate key information from images to effectively answer multi-image problems?} Previous research~\cite{shao2024visualcot, zhong2024visualtable, hu2024visualsketchpad} has demonstrated that providing useful context can enhance single-image problems and help uncover visual details that MLLMs might overlook when processing combined image and text inputs.

In this analysis, we compare different captioning strategies and derive insights into their effectiveness, focusing on four key settings: (1) concise versus detailed captions, (2) individual captions for each image versus a summarized caption across multiple images, and (3) the inclusion of questions when doing captioning. To comprehensively assess performance, we evaluate both closed-source and open-source models across all possible combinations of these factors, resulting in 8 experimental settings. For each control factor, results are averaged over the 4 relevant variations, enabling a fair and robust comparison of the different strategies.

\begin{figure*}[ht]
\centering
\includegraphics[width=0.88\textwidth]{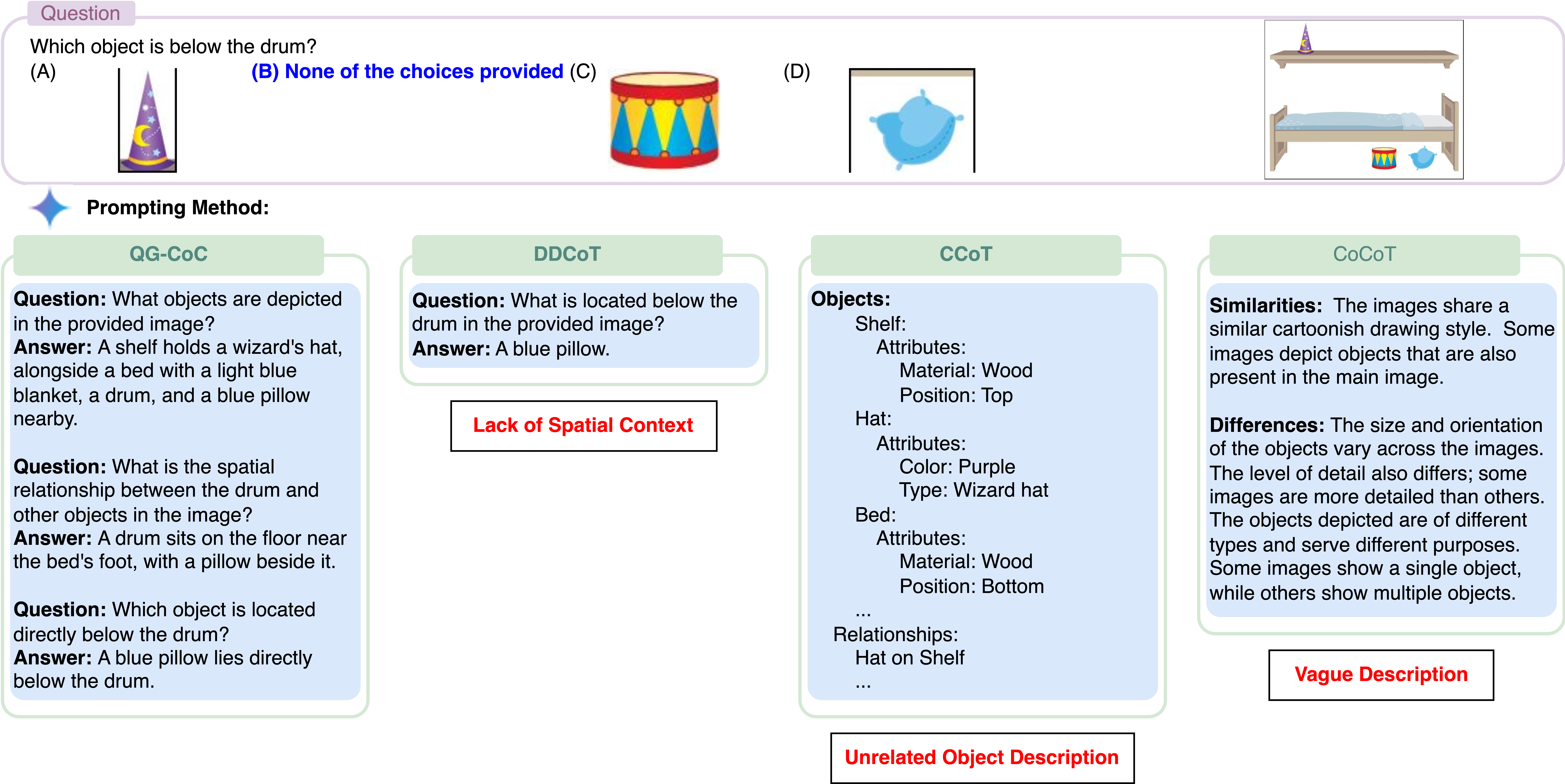}
\vspace{-2mm}
\captionsetup{font=small, width=0.88\linewidth}
\caption{An example multi-image question with different prompting methods. Text in red highlights the disadvantages. Text in blue is the correct answer. The actual prompt used for each method can be found in Appendix~\ref{sec:different_prompt}.}
\label{fig:Compare_different_prompt}
\vspace{-4mm}
\end{figure*}

% \paragraph{Different Captioning Settings Comparison.}
% We now analyze three captioning settings and provide insights based on their impact on performance:

\begin{enumerate}
    \item \textbf{Caption Length (Concise vs. Detailed)}: To examine whether the level of detail in image captions affects multi-image understanding, we compare two captioning length settings: Concise (describe the image in a sentence) vs. Detailed (describe the image in detail). Table~\ref{tab:captioning-settings} indicates that detailed captions improve multi-image accuracy due to enhanced modality matching and comprehensive image descriptions. In Figure~\ref{fig:Caption_Settings}, we can observe that detailed captioning will contain the information such as author and school list needed for answering the question. \\
    \textit{Insight}: Detailed captions are superior to concise ones, as they mitigate information loss and better support complex reasoning tasks.
    
    \item \textbf{Caption Scope (Summarized vs. Individual)}: When dealing with multiple images related to the question, a key decision is whether to summarize image set as a whole or describe each image independently. We evaluated two settings: Summarized (generate a summarized caption that describes the content across the whole set) vs. Individual (generate a separate caption for each image). Table~\ref{tab:captioning-settings} indicates that when handling multiple images, generating individual captions for each image outperforms producing a single summarized caption across all images. In Figure~\ref{fig:Caption_Settings}, we can observe that individual captioning provides more information than summarized captioning. \\
    \textit{Insight}: Individual captions are more effective than summarized captions, particularly in multi-image scenarios requiring precise, image-specific information.
    
    \item \textbf{Question-Guided (No vs. Yes)}: To understand whether integrating the question during the caption generation influences the performance, we compare two captioning settings: No Question-Guided (captions are generated based on images solely) vs. Question-Guided (captions are generated based on images and the question). Table~\ref{tab:captioning-settings} and Figure~\ref{fig:Caption_Settings} show that question-guided captions improve overall multi-image task accuracy, focusing on task-relevant visual elements. \\
    \textit{Insight}: Question-guided captioning outperforms unguided captioning by aligning generated context more closely with the question.
    
    % \item \textbf{Question Decomposition (No vs. Yes)}: For complex questions, decomposing the original query into simpler sub-questions facilitates more accurate captioning and reasoning. Table~\ref{tab:captioning-settings} observes that question decomposition improves overall multi-image task accuracy, especially for the domains that involve complex reasoning steps such as spatial and temporal reasoning. \textit{Insight}: The question-decomposition step enhances accuracy by providing structured subgoals, leading to better captioning and answers.
\end{enumerate}
% \paragraph{Insights and Implications}
% The findings from these settings indicate that nuanced captioning strategies significantly enhance MLLMs performance across multi-image tasks. Detailed captions provide a robust foundation by capturing a broader range of visual information. Individual captions excel in multi-image contexts by retaining specificity, while question-guided captioning ensures relevance to the task at hand. Finally, question decomposition offers a structured approach to tackling complexity, improving both caption quality and final answers. 
 Based on the above findings regarding effective image captioning in multi-image scenarios, the next subsection examines if adjusting the previous single-image prompting methods to multi-image scenarios can provide the necessary context for multi-image problems.

\subsection{Adjusting Existing Prompting Methods to Multi-Image Scenarios}

We conduct the following study to verify whether existing prompting methods can be effectively extended to address the complexities of multi-image scenarios.
%, we initiated an investigation into their adaptability. 
Our study focused on prominent methods such as DDCoT (Duty-Distinct Chain-of-Thought)~\cite{zheng2023ddcot}, which we adapted to decompose a central question into sub-questions applicable across multiple images; CCoT (Compositional Chain-of-Thought)~\cite{mitra2024ccot}, explored for its potential to generate a composite scene graph from each given image; and CoCoT (Contrastive Chain-of-Thought)~\cite{zhang2024cocot}, which, while originally designed for discerning similarities and differences between just two images, we considered for its conceptual applicability to broader multi-image comparisons.
As illustrated in Figure~\ref{fig:Compare_different_prompt} using Gemini-1.5-Flash~\cite{team2024gemini}, we present a case study and reveal a consistent pattern. While these adapted existing methods demonstrate some capability in identifying individual entities, their characteristics, and straightforward, explicit relationships between images, they exhibit significant limitations. Specifically, they struggle to extract deeper, implicit context or perform complex reasoning that requires synthesizing information from an arbitrary number of images. For example, DDCoT lacks present spatial context from images, CCoT presents unrelated object descriptions since it does not understand what information is needed to answer the question, and CoCoT only vaguely describes the similarity and difference between images. To further validate these observations, Section~\ref{sec:experiment} provides quantitative support that demonstrates these limitations.

Thus, since the above study highlights the need for more specialized prompting methods tailored to multi-image context, we propose a new zero-shot prompting method \textbf{Question-Guided Chain-of-Captions} that involves balancing detail, specificity, and relevance.

\begin{figure*}[t]
\centering
\includegraphics[width=0.88\textwidth]{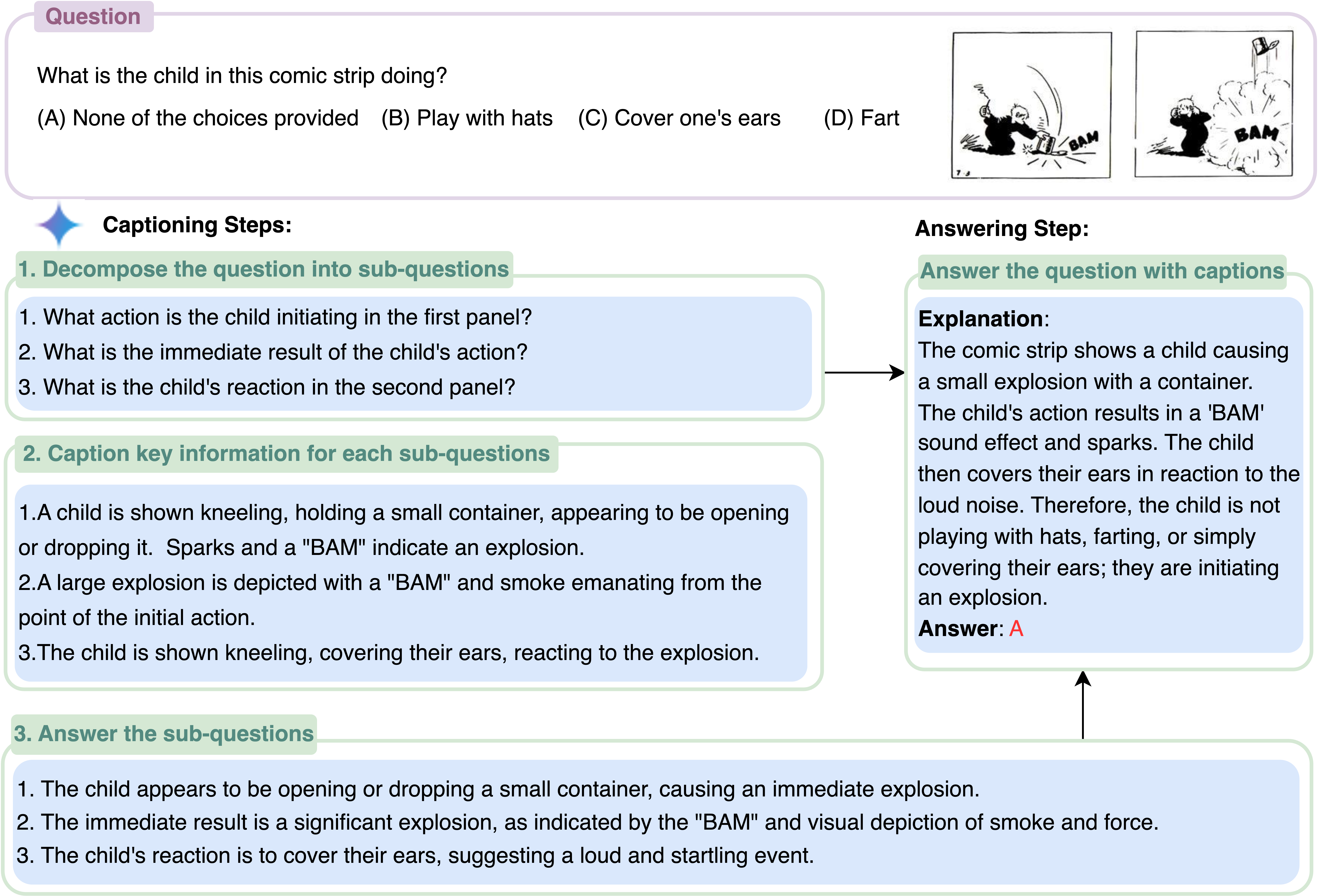}
\vspace{-2mm}
\captionsetup{font=small, width=0.9\linewidth}
\caption{An example multi-image question and its corresponding reasoning steps using QG-CoC. The prompts used for each step can be found Appendix~\ref{sec:qgcoc}.}
\label{fig:QG-CoT_pipeline}
\vspace{-4mm}
\end{figure*}

\begin{table*}[ht!]
    \centering
    \resizebox{0.9\textwidth}{!}{%
    \begin{tabular}{c c | c c c c c}
        \toprule
        \multirow{3}{*}{\textbf{Model}} & \multirow{3}{*}{\textbf{Method}} & \multicolumn{5}{c}{\textbf{Dataset}} \\
        \cmidrule(lr){3-7}
        & & \textbf{MUIR} & \textbf{MMIU} & \textbf{ScienceQA} & \textbf{MMMU} & \textbf{MMBench} \\
        \toprule
        \multicolumn{7}{c}{\textit{Open-Source}} \\
        \midrule
        % --- Existing models already here (LLaVA, Mantis) ---
        \multirow{7}{*}{LLaVA-One-Vision} 
         & w/o prompt   & 41.2 & 44.6 & \textbf{94.5} & 45.4 & 85.1 \\
         & Caption      & 42.0 {\footnotesize(+0.8)} & 48.1 {\footnotesize(+3.5)} & 91.7 {\footnotesize(-2.8)} & \textbf{49.7} {\footnotesize(+4.3)} & 85.1 {\footnotesize(+0.0)} \\
         & QG-Caption   & 44.7 {\footnotesize(+3.5)} & 49.4 {\footnotesize(+4.8)} & 93.1 {\footnotesize(-1.4)} & 45.4 {\footnotesize(+0.0)} & 85.6 {\footnotesize(+0.5)} \\
         & DDCoT        & \textbf{53.4} {\footnotesize(+12.2)} & \underline{50.5} {\footnotesize(+5.9)} & 92.9 {\footnotesize(-1.6)} & \textbf{49.7} {\footnotesize(+4.3)} & 84.3 {\footnotesize(-0.8)} \\
         & CCoT         & 44.6 {\footnotesize(+3.4)} & 46.9 {\footnotesize(+2.3)} & 93.0 {\footnotesize(-1.5)} & 46.8 {\footnotesize(+1.4)} & \underline{86.0} {\footnotesize(+0.9)} \\
         & CoCoT        & 44.2 {\footnotesize(+3.0)} & 46.4 {\footnotesize(+1.8)} & -- & -- & -- \\
         & QG-CoC       & \underline{53.3} {\footnotesize(+12.1)} & \textbf{50.9} {\footnotesize(+6.3)} & \textbf{94.5} {\footnotesize(+0.0)} & \underline{48.9} {\footnotesize(+3.5)} & \textbf{87.6} {\footnotesize(+2.5)} \\
        \midrule
        \multirow{7}{*}{Mantis-idefics2}
         & w/o prompt   & 43.4 & 45.0 & 80.3 & 41.8 & 79.0 \\
         & Caption      & 43.9 {\footnotesize(+0.5)} & 46.7 {\footnotesize(+1.7)} & 79.7 {\footnotesize(-0.6)} & 44.7 {\footnotesize(+2.9)} & 80.4 {\footnotesize(+1.4)} \\
         & QG-Caption   & 44.5 {\footnotesize(+1.1)} & 47.7 {\footnotesize(+2.7)} & 79.1 {\footnotesize(-1.2)} & 44.0 {\footnotesize(+2.2)} & 79.7 {\footnotesize(+0.7)} \\
         & DDCoT        & \underline{47.9} {\footnotesize(+4.5)} & \textbf{50.1} {\footnotesize(+5.1)} & \underline{83.0} {\footnotesize(+2.7)} & \textbf{49.7} {\footnotesize(+7.9)} & 78.3 {\footnotesize(-0.7)} \\
         & CCoT         & 44.4 {\footnotesize(+1.0)} & 44.9 {\footnotesize(-0.1)} & 80.7 {\footnotesize(+0.4)} & 46.1 {\footnotesize(+4.3)} & \underline{82.1} {\footnotesize(+3.1)} \\
         & CoCoT        & 42.6 {\footnotesize(-0.8)} & 45.4 {\footnotesize(+0.4)} & -- & -- & -- \\
         & QG-CoC       & \textbf{48.9} {\footnotesize(+5.5)} & \underline{49.8} {\footnotesize(+4.8)} & \textbf{83.8} {\footnotesize(+3.5)} & \underline{48.9} {\footnotesize(+7.1)} & \textbf{83.4} {\footnotesize(+4.4)} \\
        \midrule
        \multirow{7}{*}{Qwen-2.5-VL}
         & w/o prompt   & 62.1 & 50.3 & 90.2 & 58.2 & 88.2 \\
         & Caption      & 62.8 {\footnotesize(+0.7)} & 50.9 {\footnotesize(+0.6)} & 88.0 {\footnotesize(-2.2)} & 59.4 {\footnotesize(+1.2)} & 88.3 {\footnotesize(+0.1)} \\
         & QG-Caption   & 62.4 {\footnotesize(+0.3)} & 50.1 {\footnotesize(-0.2)} & 88.9 {\footnotesize(-1.3)} & 60.0 {\footnotesize(+1.8)} & 88.5 {\footnotesize(+0.3)} \\
         & DDCoT        & 63.7 {\footnotesize(+1.6)} & 54.1 {\footnotesize(+3.8)} & 90.5 {\footnotesize(+0.3)} & 61.5 {\footnotesize(+3.3)} & 87.9 {\footnotesize(-0.3)} \\
         & CCoT         & 62.3 {\footnotesize(+0.2)} & 51.6 {\footnotesize(+1.3)} & 89.5 {\footnotesize(-0.7)} & 59.5 {\footnotesize(+1.3)} & 88.5 {\footnotesize(+0.3)} \\
         & CoCoT        & 62.6 {\footnotesize(+0.5)} & 52.3 {\footnotesize(+2.0)} & -- & -- & -- \\
         & \textbf{QG-CoC} & \textbf{65.3} {\footnotesize(+3.2)} & \textbf{56.9} {\footnotesize(+6.6)} & \textbf{91.9} {\footnotesize(+1.7)} & \textbf{64.8} {\footnotesize(+6.6)} & \textbf{89.4} {\footnotesize(+1.2)} \\
        \midrule
        \multicolumn{7}{c}{\textit{Closed-Source}} \\
        \midrule
        \multirow{7}{*}{GPT-4o}
         & w/o prompt   & 70.8 & 63.3 & 89.5 & 63.1 & 86.0 \\
         & Caption      & 71.8 {\footnotesize(+1.0)} & 63.6 {\footnotesize(+0.3)} & 86.8 {\footnotesize(-2.7)} & \underline{66.0} {\footnotesize(+2.9)} & 88.1 {\footnotesize(+2.1)} \\
         & QG-Caption   & 70.0 {\footnotesize(-0.8)} & \underline{65.1} {\footnotesize(+1.8)} & \underline{89.6} {\footnotesize(+0.1)} & 61.7 {\footnotesize(-1.4)} & \textbf{89.5} {\footnotesize(+3.5)} \\
         & DDCoT        & 73.1 {\footnotesize(+2.3)} & 62.9 {\footnotesize(-0.4)} & 89.3 {\footnotesize(-0.2)} & 64.5 {\footnotesize(+1.4)} & 86.6 {\footnotesize(+0.6)} \\
         & CCoT         & 70.4 {\footnotesize(-0.4)} & 60.9 {\footnotesize(-2.4)} & 87.8 {\footnotesize(-1.7)} & 61.0 {\footnotesize(-2.1)} & 88.1 {\footnotesize(+2.1)} \\
         & CoCoT        & \underline{74.0} {\footnotesize(+3.2)} & 64.5 {\footnotesize(+1.2)} & -- & -- & -- \\
         & QG-CoC       & \textbf{74.9} {\footnotesize(+4.1)} & \textbf{65.8} {\footnotesize(+2.5)} & \textbf{90.3} {\footnotesize(+0.8)} & \textbf{66.7} {\footnotesize(+3.6)} & \underline{88.9} {\footnotesize(+2.9)} \\
        \midrule
        \multirow{7}{*}{Gemini-1.5-Flash}
         & w/o prompt   & 66.0 & 55.0 & \underline{87.0} & \underline{64.5} & \textbf{86.0} \\
         & Caption      & 66.8 {\footnotesize(+0.8)} & 53.7 {\footnotesize(-1.3)} & 86.9 {\footnotesize(-0.1)} & 61.0 {\footnotesize(-3.5)} & 84.5 {\footnotesize(-1.5)} \\
         & QG-Caption   & 66.0 {\footnotesize(+0.0)} & 54.9 {\footnotesize(-0.1)} & 86.8 {\footnotesize(-0.2)} & \textbf{66.7} {\footnotesize(+2.2)} & 84.9 {\footnotesize(-1.1)} \\
         & DDCoT        & \underline{67.6} {\footnotesize(+1.6)} & 51.5 {\footnotesize(-3.5)} & 86.9 {\footnotesize(-0.1)} & 53.9 {\footnotesize(-10.6)} & 84.5 {\footnotesize(-1.5)} \\
         & CCoT         & 66.3 {\footnotesize(+0.3)} & 51.9 {\footnotesize(-3.1)} & 85.5 {\footnotesize(-1.5)} & 53.2 {\footnotesize(-11.3)} & \underline{85.6} {\footnotesize(-0.4)} \\
         & CoCoT        & 65.4 {\footnotesize(-0.6)} & \textbf{55.5} {\footnotesize(+0.5)} & -- & -- & -- \\
         & QG-CoC       & \textbf{68.2} {\footnotesize(+2.2)} & \underline{55.4} {\footnotesize(+0.4)} & \textbf{87.2} {\footnotesize(+0.2)} & 63.7 {\footnotesize(-0.8)} & 85.2 {\footnotesize(-0.8)} \\
        \bottomrule
    \end{tabular}
    }
    \captionsetup{font=small, width=0.9\linewidth}
    \caption{Multi-Image and Single-Image benchmark performance of different models with various prompting methods. Numbers in {\footnotesize(+/-)} indicate delta compared to the w/o prompt baseline of the same model. Metrics represent answer accuracy (\%).}
    \label{tab:main_result}
    \vspace{-6mm}
\end{table*}

% The method first decomposes a given question into a sequence of sub-questions. Each sub-question focuses on a specific aspect of the image. Then, the model generates targeted captions that describe the key visual elements corresponding to each sub-question. By sequentially answering these sub-questions based on the captions, the method builds an interpretable chain of reasoning that culminates in answering the original question. This process ensures that the model’s interpretation remains closely aligned with the question's intent and the visual evidence, promoting greater accuracy and explainability. In particular, QG-CoC demonstrates strong performance on tasks requiring multi-step visual reasoning. We also include the actual prompts used in each step in Appendix.
\section{Question-Guided Chain-of-Captions}

As shown in Figure~\ref{fig:QG-CoT_pipeline}, Question-Guided Chain-of-Captions (QG-CoC) is a structured reasoning approach designed to enhance multi-image understanding. The method involves three key steps:
\paragraph{Step 1: Decompose the question into sub-questions.}
First, given a complex question, the method breaks it down into a series of simpler, interpretable sub-questions. Each sub-question targets a specific aspect of the image(s), such as the subject’s action, outcome, or reaction. This decomposition ensures that the reasoning is detailed and aligned with the intent of the question.
\paragraph{Step 2: Caption key information for each sub-question.}
The MLLM then generates targeted captions for each sub-question. These captions extract and describe the most relevant visual evidence (e.g., objects, actions, effects, or scene changes), providing intermediate interpretations. This step directly connects each piece of reasoning to the image content.
\paragraph{Step 3: Answer the sub-questions and integrate reasoning.}
Finally, the model answers each sub-question based on the captions, forming a coherent reasoning chain. These individual answers are then combined to produce the final answer to the original question, supported by visual evidence from the images. This step-by-step process improves both the accuracy and the explainability of the model predictions.

\section{Experimental Results}
\label{sec:experiment}
\subsection{Experimental Setting}
\paragraph{Implementation.}
We conduct experiments using different zero-shot prompting methods on both closed-source and open-source MLLMs. For experiments in this section, we utilize GPT-4o~\cite{hurst2024gpt} and Gemini-1.5-Flash~\cite{team2024gemini} as representatives of general-purpose MLLMs.
We also utilize two open-sourced MLLMs: Mantis-idefics2-8B~\cite{jiang2024mantis}, LLaVA-OneVision-7B~\cite{li2024llava}, and Qwen-2.5-VL-7B~\cite{bai2025qwen2}, which support multiple image inputs. However, they have limited capacity to process and follow long prompts to generate additional context in the first stage. From open-source model evaluation, we use Gemini-1.5-Flash as oracle captioning in the first stage. The versions of these models we used for the experiments are listed in Appendix~\ref{sec:hyperparameters}.

\paragraph{Baselines.}
First, to evaluate the added benefit of our method to pretrained MLLMs, our default baseline is to apply the model to the benchmark without any prompt engineering. Then, we compare \textbf{QG-CoC} prompting to five state-of-the-art methods including: (1) \textbf{Detailed Captioning}: In the previous section, we find that captioning image individually in detail enhance the performance the most, (2) \textbf{Question-Guided Detailed Captioning}: In the previous section, we find that adding question in the prompt enhances the performance, (3) \textbf{DDCoT}~\cite{zheng2023ddcot}: First, decompose the question, then utilizes MLLMs to answer the sub-questions and uses it as rationale, (4) \textbf{CCoT}~\cite{mitra2024ccot}: Utilize MLLMs to generate a scene graph based on each image, and (5) \textbf{CoCoT}~\cite{zhang2024cocot}: Utilize MLLMs to describe the similarity and difference between multiple images. All these methods work in a two-step pipeline. The first step generates an additional textual representation from the instructions of different methods. The second step involves passing the images, question, and output from the first step to answer the question.

\paragraph{Evaluation Dataset.}
%Since there have been many multi-image benchmarks recently, 
We select two representative and multi-faceted benchmarks: MuirBench~\cite{wang2024muirbench} and MMIU~\cite{meng2024mmiu}.
MuirBench is a comprehensive benchmark consisting of 12 diverse multi-image tasks, such as scene understanding, ordering, etc. It contains 2,600 multiple-choice questions with 11,264 images in total. We report the overall average performance across the 12 tasks.
MMIU is a multi-image benchmark encompassing 7 types of multi-image relationships, 52 tasks, 77K images, and 11K multiple-choice questions. We report the overall average performance across all the tasks. However, during the evaluation, we observe some tasks in MMIU exhibit low quality, so we filter out some tasks in the spatial and semantic relationships. We also compare our method on various single-image tasks, including MMMU~\cite{yue2024mmmu}, MMBench~\cite{liu2024mmbench}, and ScienceQA~\cite{lu2022scienceqa}, to validate the generalizability of our method. However, since CoCoT is constructed under image comparison, we cannot evaluate CoCoT on single-image benchmarks.

\subsection{Main Results}
% MMIU MUIR (GPT-4o, Gemini-Flash, Mantis, LLava-OV)
% MMMU MMBench ScienceQA
To investigate which prompting methods and models better solve multi-image problems, we summarize the answer accuracy performance in Table~\ref{tab:main_result}.

% \paragraph{Comparison with various prompting baselines.}
% QG-CoC demonstrates strong performance compared to various prompting baselines under both multi-image and single-image benchmarks. In the previous section, we observe that providing detailed captions for individual images boosts performance, as does incorporating the question into the prompt (Question-Guided Detailed Captioning), especially under the multi-image benchmark.
% DDCoT sometimes shows competitive performance, but QG-CoC often outperforms it on the multi-image benchmarks. CCoT and CoCoT show generally much lower accuracy than QG-CoC on the multi-image benchmarks. This indicates that generating a scene graph and describing the image similarity and difference might not be as directly beneficial for general multi-image reasoning. Open-source models gain a larger performance gap from our method in multi-image benchmarks, which means that without training with a large amount of multi-image data, we can still improve the performance by providing useful content.
% Thus, the results indicate the effectiveness of QG-CoC in leveraging both detailed image understanding and question-aware reasoning.

\paragraph{Comparison with various prompting baselines.}
QG-CoC demonstrates strong performance across both multi-image and single-image benchmarks, as shown in Table~\ref{tab:main_result}:
\begin{enumerate}
    \item \textbf{Comparison over Caption:} While providing detailed captions for individual images (``Caption'' method) is beneficial, QG-CoC not only provides image captions but also ensures these captions are directly relevant to specific parts of the sub-question. This relevance is achieved by first decomposing the main question into sub-questions (\textit{Step 1}) and captioning key information for sub-questions (\textit{Step 2}). As a result, the generated captions are targeted, leading to more focused and effective reasoning compared to general detailed captions. 
    %(e.g., for LLaVA-One-Vision on MMIU, QG-CoC: 50.9 vs Caption: 48.1 and GPT-4o on MUIR, QG-CoC: 74.9 vs Caption: 71.8).

    \item \textbf{Comparison over QG-Caption:} QG-Caption incorporates the question into the prompt to improve caption relevance. Instead of guiding captions with a single, potentially complex main question, QG-CoC decomposes the question into simpler sub-questions (\textit{Step 1}) and then generates targeted captions for each sub-question (\textit{Step 2}). This question-guided captioning at each sub-question typically yields better results than a single pass of QG-Caption. 
    %(e.g., for LLaVA-One-Vision on MMIU, QG-CoC: 50.9 vs QG-Caption: 49.4 and GPT-4o on MMIU, QG-CoC: 65.8 vs QG-Caption: 65.1).

    \item \textbf{Comparison over DDCoT:} DDCoT also involves question decomposition. However, QG-CoC introduces a crucial intermediate step: generating explicit, targeted captions for each sub-question (\textit{Step 2}) before proceeding to answer them and integrate reasoning (\textit{Step 3}). This step of grounding each sub-problem in visual evidence through dedicated captions often leads to more robust reasoning. While DDCoT shows competitive performance, QG-CoC frequently outperforms it. 
    %(e.g., for LLaVA-One-Vision on MMIU, QG-CoC: 50.9 vs. DDCoT: 50.5 and GPT-4o on MUIR, QG-CoC: 74.9 vs. DDCoT: 73.1).

    \item \textbf{Comparison over CCoT:} While scene graphs can be informative, they might produce overly detailed or less relevant information for a specific question. Our method of generating captions related to sub-questions (\textit{Step 2}), guided by the initial question decomposition (\textit{Step 1}), ensures that the visual information extracted is directly relevant to the task. Thus, QG-CoC consistently demonstrates higher accuracy than CCoT. 
    %(e.g., for LLaVA-One-Vision on MMIU, QG-CoC: 50.9 vs. CCoT: 46.9; for GPT-4o on MMIU, QG-CoC: 65.8 vs. CCoT: 60.9).

    \item \textbf{Comparison over CoCoT:} CoCoT utilizes MLLMs to describe the similarity and difference between multiple images. This can be effective for comparative tasks but may not be optimal for all types of multi-image tasks. QG-CoC, through its sub-question decomposition (\textit{Step 1}) and subsequent targeted captioning (\textit{Step 2}), offers a more general framework that can adapt to various reasoning needs beyond simple comparison. As a result, QG-CoC generally achieves higher accuracy than CoCoT. 
    %(e.g., for LLaVA-One-Vision on MMIU, QG-CoC: 50.9 vs. CoCoT: 46.4; for Mantis-idefics2 on MUIR, QG-CoC: 48.9 vs. CoCoT: 42.6).
\end{enumerate}
% In particular, open-source models gain a larger performance improvement from QG-CoC method in multi-image benchmarks, as seen in Table~\ref{tab:main_result}. This suggests that even without training on excessive multi-image data, performance can be substantially enhanced by providing structured and useful visual content through QG-CoC's steps. 
Overall, the results show the effectiveness of QG-CoC in leveraging both detailed image understanding and question-aware reasoning.
\section{Discussion}

We conduct an analysis of QG-CoC through multiple perspectives, including detailed breakdowns of different visual domains on MMIU and MUIR benchmarks, the impact of incorporating each component of QG-CoC, and common error analysis.

% \paragraph{Human Validation on Caption Method.}

\begin{figure*}[ht]
 \centering
  % \begin{subfigure}{0.32\textwidth}
  %   \includegraphics[width=\linewidth]{Figure/gemini_MMIU_each_dim.png}
  %   \captionsetup{font=small, width=0.8\linewidth}
  %   \caption{Gemini-1.5-Flash}
  %   \label{fig:all_dim_1}
  % \end{subfigure}
  % \hfill % This command adds a horizontal space between the subfigures
  \begin{subfigure}{0.48\textwidth}
    \includegraphics[width=\linewidth]{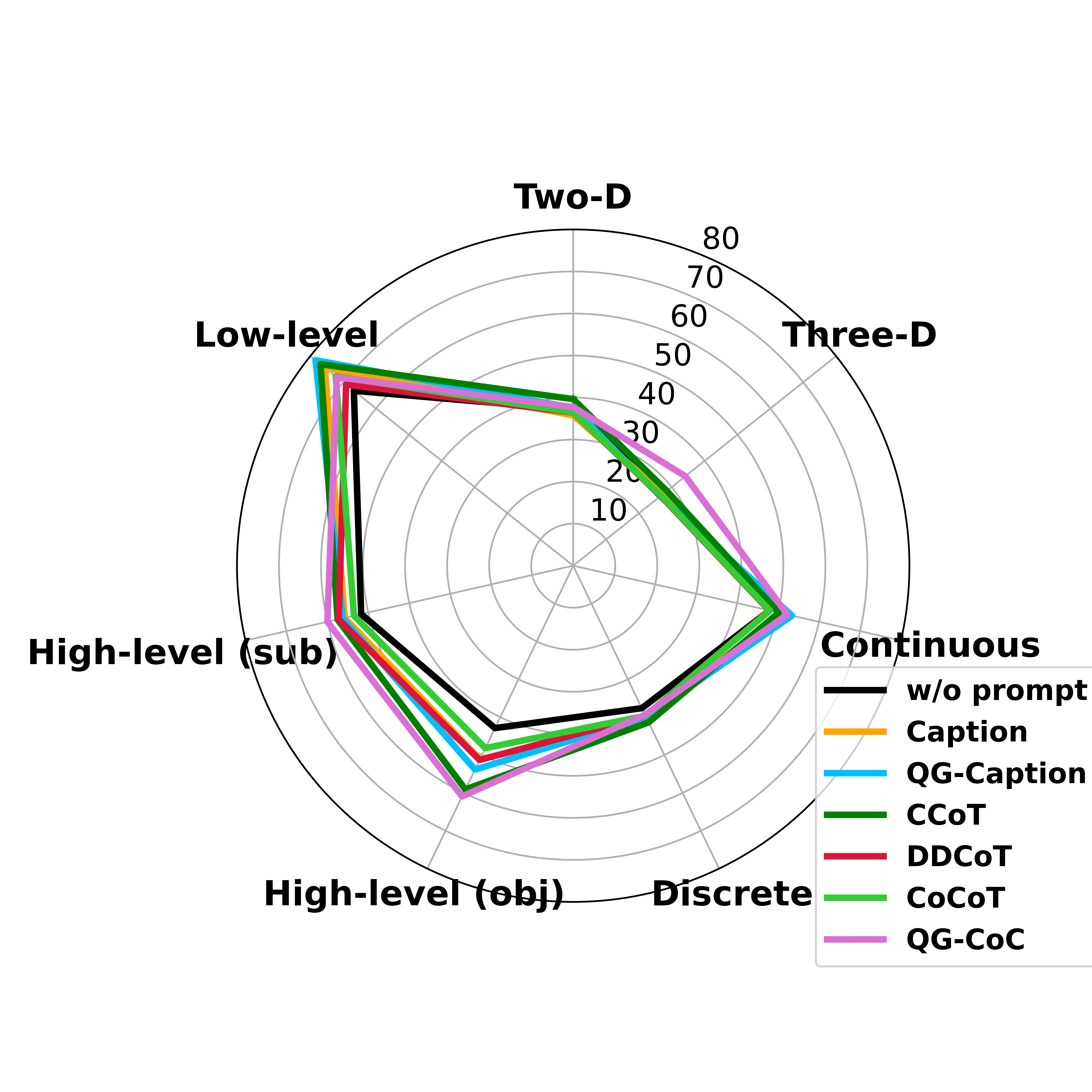}
    \captionsetup{font=small, width=0.8\linewidth}
    \vspace{-4mm}
    \caption{LLaVA-OV}
    \label{fig:all_dim_2}
  \end{subfigure}
\hspace{0.02\textwidth}
  \begin{subfigure}{0.48\textwidth}
    \includegraphics[width=\linewidth]{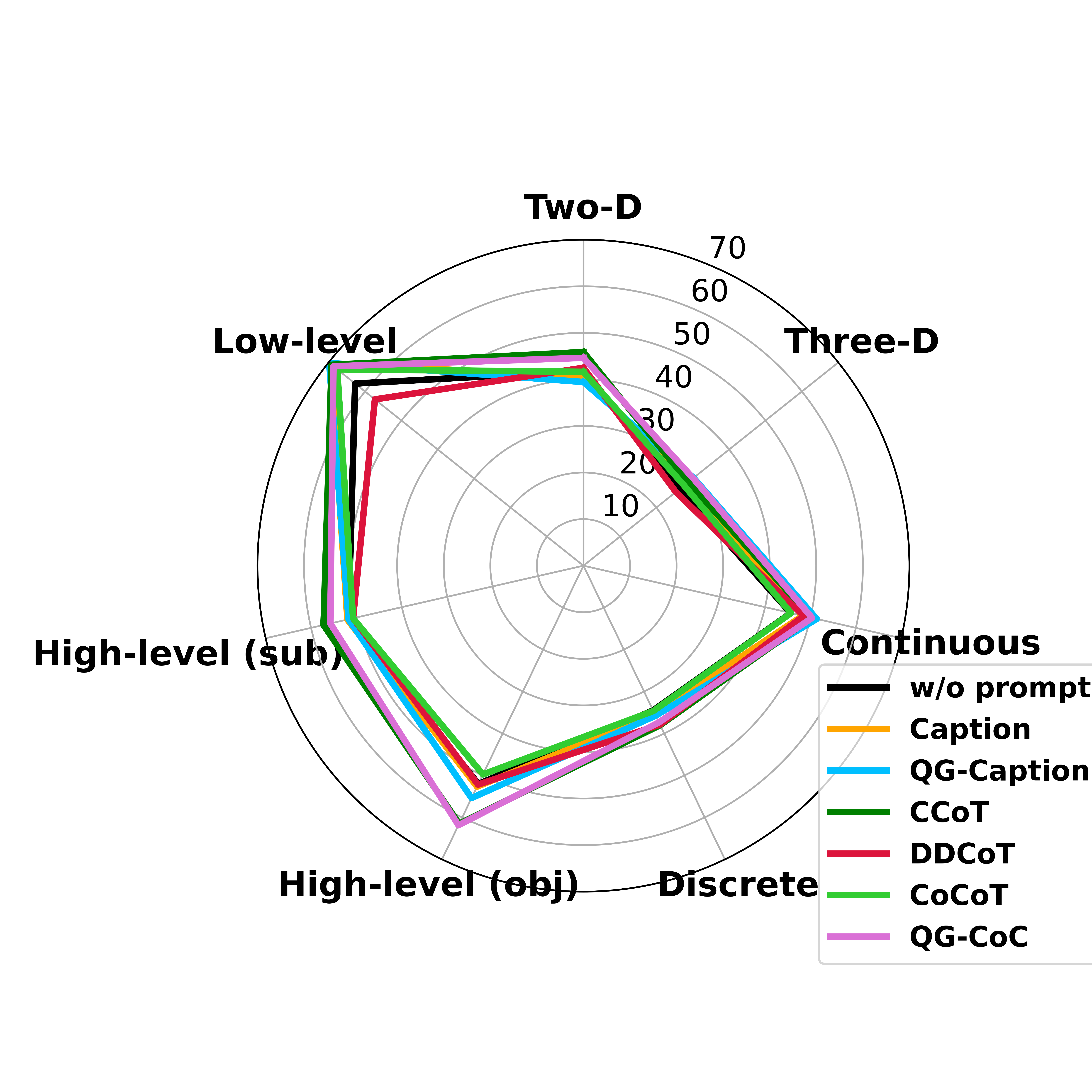}
    \vspace{-4mm}
    \captionsetup{font=small, width=0.8\linewidth}
    \caption{Mantis}
    \label{fig:all_dim_3}
  \end{subfigure}
\captionsetup{font=small, width=0.9\linewidth}
 \caption{Prompting methods performance by image relationships on different models (MMIU dataset).}
 \label{fig:examples_all_dim_plot}
 \vspace{-4mm}
\end{figure*}

\paragraph{Different Prompting Methods Performance Across Various Image Relationships.}

 As shown in Figure~\ref{fig:examples_all_dim_plot}, models exhibit different capabilities across various image relationships in MMIU. 
 % Specifically, different prompting methods excel at understanding semantic content in multi-image scenarios, perform moderately in temporal tasks, and obtain the worst performance in spatial tasks. 
 We also record all model performance on all tasks in MMIU (Table~\ref{tab:mmiu_prompting_comparison_detailed}) and MUIR (Table~\ref{tab:muir_prompting_comparison_detailed}).

1) In semantic relationships, direct prompting generally performs better on multi-image semantic tasks involving low-level relationships than adding more context. Since low-level relationships usually involve intuitive understanding, providing more details will not help with reasoning. Inversely, in high-level tasks, for subjective tasks such as Causality Reasoning and Emotion Recognition, which require the identification and reasoning of implicit visual information, and objective tasks, such as retrieval tasks, QG-CoC outperforms existing methods significantly since our method provides more key information to tackle them. 2) In temporal relationships, all prompting methods can handle discrete and continuous temporal relationships relatively well, but perform poorly on reasoning-intensive tasks such as Visual Ordering and Temporal Ordering. 3) In spatial relationships, we find that all prompting method struggles with understanding both 2D and 3D positional relations. Since these prompting methods cannot provide spatial information in multiple images and reason correctly, QG-CoC overall provides more spatial-related information compared to other methods.

% \begin{table*}[h]
% \centering
% \small
% \resizebox{14cm}{!}{%
% \begin{tabular}{ccccccccc}
% \thickhline
% \textbf{Question-Decompose} & \textbf{Question-Aware Caption} & \textbf{Solver} & \textbf{Equation} & \textbf{MUIR}      & \textbf{MMIU}      \\ \hline
% \xmark           & \xmark          & \cmark      & \cmark        & 77.8\% & 74.0\% \\
% \xmark           & \cmark          & \cmark      & \cmark        & \underline{79.6\%} \\
% \cmark           & \xmark          & \cmark      & \cmark        & 64.1\% & \underline{81.2\%}  \\
% \cmark           & \cmark          & \xmark      & \cmark        & 59.3\% & 73.3\% \\
% \cmark           & \cmark          & \cmark      & \cmark        & \textbf{85.2\%} & \textbf{87.6\%} \\ \thickhline
% \end{tabular}}
% \label{tab:ablation}
% \end{table*}

\begin{table}[ht]
\centering
\begin{adjustbox}{width=0.8\linewidth}
\begin{tabular}{lcc}
\toprule
\textbf{Method} & \textbf{MUIR} & \textbf{MMIU} \\
\midrule
Zero-shot & 66.0 & 55.0 \\
+ Question-Decompose & 66.5 & 54.8 \\
+ Question-Guided Caption & 67.2 & 55.1 \\
+ QG-CoC & \textbf{68.2} & \textbf{55.4} \\
\bottomrule
\end{tabular}
\end{adjustbox}
\vspace{-2mm}
\captionsetup{font=small, width=0.9\linewidth}
\caption{Ablation experiment results across MMIU and MUIR benchmarks using Gemini-1.5-Flash. Our method achieves the highest accuracy among all.}
\label{tab:ablation}
\vspace{-6mm}
\end{table}

\paragraph{Importance of each component on QG-CoC.}
We analyze the contribution of each component in QG-CoC through an ablation study on the MUIR and MMIU benchmarks. In Table~\ref{tab:ablation}, starting from the zero-shot baseline, each successive module leads to consistent performance gains. Introducing Question Decomposition improves MUIR accuracy from 66.0 to 66.5, showing the benefit of simplifying complex queries. Adding the Question-Guided Captioning module further raises the score to 67.2, highlighting the importance of context-aware visual grounding. Finally, incorporating the full QG-CoC model achieves the highest accuracy of 68.2 on MUIR and 55.4 on MMIU, confirming that the combined reasoning and generation steps effectively enhance overall understanding. These results underscore the complementary roles of each module and validate the design of our compositional reasoning pipeline.
                        % MMIU            MUIR
% Zero-shot
% + Question-Decompose 
% + Question-Aware Caption
% + QG-CoT

% \subsection{Our Captioning + Baseline Methods}
% Gemini-Flash	MMIU	MUIR \\
% Scene Graph	\\	
% plus Ours	\\		
% Duty Decompose \\		
% plus Ours		\\
% Similarities and Differences	\\
% plus Ours		\\

\begin{table}[ht!]
\centering
\small
\resizebox{7cm}{!}{%
\begin{tabular}{lc}
\thickhline
\textbf{Error Reason} & \textbf{Percentage (\%)} \\ \hline
(E1) Wrong question understanding & 33.3\% (40/120) \\ 
(E2) Inaccurate perception & 31.7\% (38/120) \\ 
(E3) Wrong reasoning & 35.0\% (42/120) \\ \thickhline
\end{tabular}}
\vspace{-2mm}
\captionsetup{font=small, width=0.9\linewidth}
\caption{Statistics of error analysis under Gemini-1.5-Flash using QG-CoC.}
\vspace{-6mm}
\label{tab:error_analysis_table}
\end{table}
\begin{table*}[ht!]
    \centering
    \resizebox{\textwidth}{!}{%
    \begin{tabular}{lcccccccccccc}
        \toprule
        \textbf{Error Type} & \textbf{Geographic} & \textbf{Diagram} & \textbf{Matching} & \textbf{Difference} & \textbf{Retrieval} & \textbf{Counting} & \textbf{Attribute} & \textbf{Scene} & \textbf{Action} & \textbf{Grounding} & \textbf{Cartoon} & \textbf{Ordering} \\
        \midrule
        \textbf{E1} & 50 & 30 & 30 & 30 & 30 & 40 & 30 & 30 & 30 & 30 & 30 & 40 \\
        \textbf{E2} & 30 & 40 & 40 & 20 & 40 & 40 & 40 & 20 & 20 & 30 & 30 & 30 \\
        \textbf{E3} & 20 & 30 & 30 & 50 & 30 & 20 & 30 & 50 & 50 & 40 & 40 & 30 \\
        \bottomrule
    \end{tabular}}
    \caption{Distribution of error types (\%) across MUIR tasks for Gemini-1.5-Flash under QG-CoC prompting.}
    \label{tab:error_analysis_task}
\vspace{-6mm}
\end{table*}

\paragraph{Error Analysis.}
We delve deeper into the primary challenges that MLLMs encounter when solving multi-image problems using QG-CoC. To gain a quantitative understanding of model failures, we randomly sample 10 error instances for every task and a total of 120 error instances made by Gemini-1.5-flash on MuirBench, and annotate the main reasons for these mispredictions.
We categorize into the three error types, including: \textbf{(E1) Wrong question understanding}, which means MLLMs do not understand the question accurately, leading to the incorrect question decomposition. \textbf{(E2)
Wrong perception}, which means the failure to capture details in or between images. \textbf{(E3) Wrong reasoning}, which means even if we get accurate decomposition and captioning, MLLMs still infer the wrong reasoning path to answer the question.

In Table~\ref{tab:error_analysis_table}, we observe that the most common error category (35.0\% of error cases) is failure of reasoning. We conclude that even if the given context is accurate, MLLMs still infer incorrectly. The other error category (33.3\% of error cases) is due to inaccurate question understanding and influences the generation of incorrect captions and reasoning. The rest 31.7\% of errors are due to the failure to capture details in images. The detailed qualitative examples are provided in Figure~\ref{fig:error_cases}.

We further analyze errors by task category in MUIR (Table~\ref{tab:error_analysis_task}). We observe that tasks requiring holistic multi-image understanding (e.g., Difference, Scene, Action) are dominated by E3. In contrast, tasks relying on fine-grained perception (e.g., Matching, Attribute, Counting) are more prone to E2. Meanwhile, E1 is consistently present, with higher prevalence in abstract tasks like Ordering and Geographic. Overall, the breakdown confirms that reasoning across multiple images remains the most significant challenge.

\begin{table}[ht!]
\centering
\small
\begin{adjustbox}{width=0.8\linewidth}
\begin{tabular}{lcc}
\hline
\textbf{Method} & \textbf{\#Tokens} & \textbf{Runtime} \\
\hline
w/o prompt & 0 & 3.5s \\
Caption & 349 & 8.5s \\
QG-Caption & 169 & 6.6s \\
DDCoT & 108 & 5.8s \\
CCoT & 372 & 8.7s \\
CoCoT & 111 & 5.9s \\
\textbf{QG-CoC} & \textbf{127} & \textbf{6.1s} \\
\hline
\end{tabular}
\end{adjustbox}
\captionsetup{font=small, width=0.9\linewidth}
\caption{Computational Overhead Analysis on MMIU Benchmark. Runtime means the average runtime(seconds) per sample. \#Tokens means the average additional tokens per sample.}
\label{tab:computational_overhead}
\vspace{-6mm}
\end{table}

\paragraph{Inference Time Comparison Analysis.}
We analyze the computational overhead of our method, QG-CoC. The method involves a two-stage pipeline, which inherently introduces additional costs compared to direct prompting. To quantify this, we measured the extra token usage for closed-source models, using Gemini-1.5-Flash as an example, and the inference runtime for open-source models, exemplified by LLaVA-OneVision-7B. The results, averaged on 100 data samples randomly selected from the MMIU benchmark and run on 4 NVIDIA A6000 GPUs for open-source models, are detailed in Table~\ref{tab:computational_overhead}. For Gemini-1.5-Flash, token estimation was based on the Google-provided API. As the table indicates, QG-CoC does increase token usage and runtime. However, we contend that this is a justifiable trade-off for the consistent performance improvements documented in our paper. This is particularly evident for open-source models, where QG-CoC leads to more significant gains, with a +12\% improvement for LLaVA-OV and +5\% for Mantis. The overhead is comparable to other two-stage methods while achieving superior accuracy. We believe this represents an efficient utilization of resources to unlock more advanced reasoning capabilities.

\section{Conclusion}
In this work, we introduce a novel prompting method called Question-Guided Chain-of-Captions (QG-CoC), which first incorporates problem decomposition and then generates each sub-question-guided image captioning to provide a clue to answer the sub-question, then combines the sub-question and sub-answer pair as prior knowledge to answer the original problem. Our extensive experiments demonstrate the advantages of our method for different MLLMs on various benchmarks.
\section*{Limitations}
This work only provides a strong baseline for the single-image and multi-image reasoning of MLLMs. Although we experiment with many representative models and reasoning methods in this paper, we acknowledge that this does not cover all models and frameworks. Our proposed method relies on the captioning ability of advanced MLLMs. Therefore, it might cause performance deterioration in less advanced language models or more challenging tasks. 
To strengthen QG-CoC, a more diverse and complicated scenario should be explored in the future, such as complex geometric shapes and even 2D, 3D-spatial information.

% \section*{Acknowledgments}

% This document has been adapted
% by Steven Bethard, Ryan Cotterell and Rui Yan
% from the instructions for earlier ACL and NAACL proceedings, including those for
% ACL 2019 by Douwe Kiela and Ivan Vuli\'{c},
% NAACL 2019 by Stephanie Lukin and Alla Roskovskaya,
% ACL 2018 by Shay Cohen, Kevin Gimpel, and Wei Lu,
% NAACL 2018 by Margaret Mitchell and Stephanie Lukin,
% Bib\TeX{} suggestions for (NA)ACL 2017/2018 from Jason Eisner,
% ACL 2017 by Dan Gildea and Min-Yen Kan,
% NAACL 2017 by Margaret Mitchell,
% ACL 2012 by Maggie Li and Michael White,
% ACL 2010 by Jing-Shin Chang and Philipp Koehn,
% ACL 2008 by Johanna D. Moore, Simone Teufel, James Allan, and Sadaoki Furui,
% ACL 2005 by Hwee Tou Ng and Kemal Oflazer,
% ACL 2002 by Eugene Charniak and Dekang Lin,
% and earlier ACL and EACL formats written by several people, including
% John Chen, Henry S. Thompson and Donald Walker.
% Additional elements were taken from the formatting instructions of the \emph{International Joint Conference on Artificial Intelligence} and the \emph{Conference on Computer Vision and Pattern Recognition}.

\bibliography{custom}

\appendix

\section{Model Hyperparameters}
\label{sec:hyperparameters}

\begin{figure*}[h]
 \centering
  \begin{subfigure}{0.48\textwidth}
    \includegraphics[width=\linewidth]{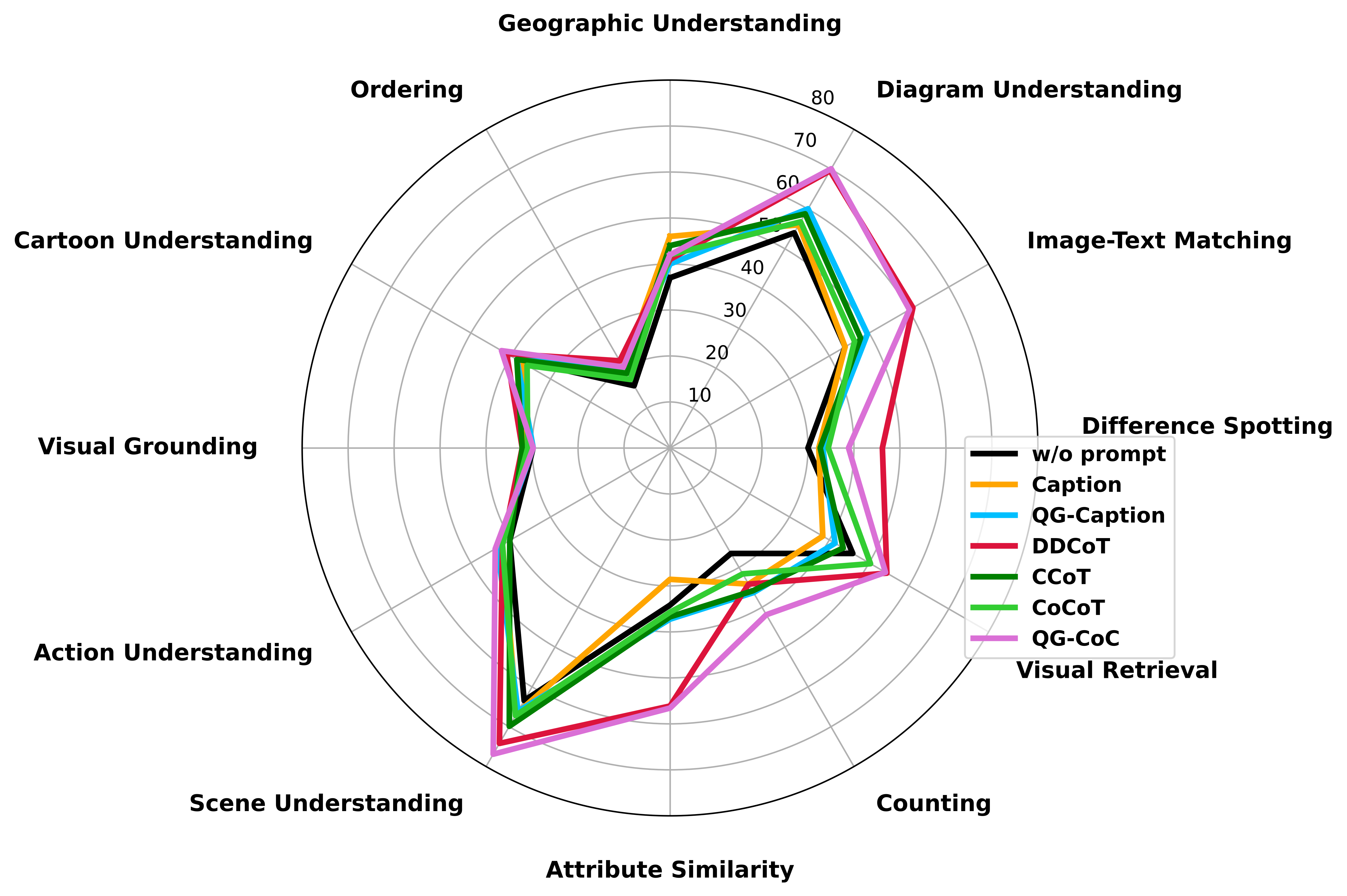}
    \captionsetup{font=small, width=0.8\linewidth}
    \caption{LLaVA-OV}
    \label{fig:muir_all_dim_2}
  \end{subfigure}
  \hfill
  \begin{subfigure}{0.48\textwidth}
    \includegraphics[width=\linewidth]{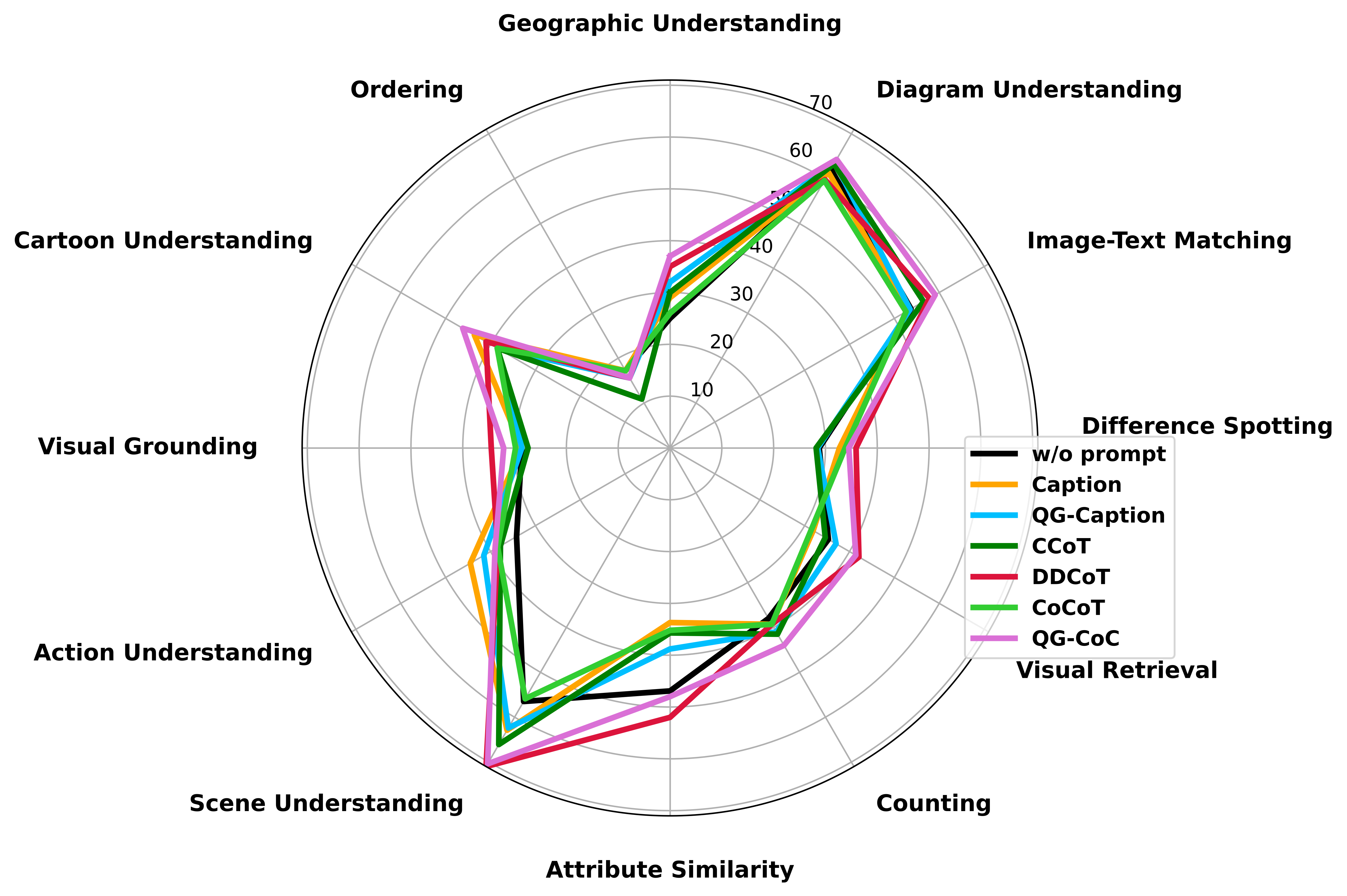}
    \captionsetup{font=small, width=0.8\linewidth}
    \caption{Mantis}
    \label{fig:muir_all_dim_3}
  \end{subfigure}
\captionsetup{font=small, width=0.9\linewidth}
 \caption{Prompting methods performance by tasks on different models. (MUIR)}
 \label{fig:examples_muir_all_dim_plot}
\end{figure*}

\begin{table*}[ht]
\centering
\small
\resizebox{16cm}{!}{
\begin{tabular}{lll}
\thickhline
Model                 & Version                                          & Generation Setup                                      \\ \hline
\multicolumn{3}{c}{\textit{Close-source}}                                                                                                 \\ \hline
GPT-4o               & gpt-4o-2024-05-13                               & temperature = 0, max tokens = 2048                    \\
Gemini-Flash            & gemini-1.5-flash                                   & temperature = 0, max tokens = 2048                    \\ \hline
\multicolumn{3}{c}{\textit{Open-source}}                                                                                                  \\ \hline
LLaVA-OneVision-7B   & lmms-lab/llava-onevision-qwen2-7b-ov                          & do\_sample=False, temperature=0, max tokens = 2048 \\
Mantis-Idefics2-8B   & TIGER-Lab/Mantis-8B-Idefics2               & do\_sample=False, temperature=0, max tokens = 2048 \\ \thickhline
\end{tabular}}
\captionsetup{font=small, width=0.9\linewidth}
\caption{Model names, versions, and generating setups for various MLLMs.}
\label{tab:hyperparameter}
\end{table*}
The hyperparameters for the experiments for studying QG-CoC and other prompting methods are set to their default values to ensure consistency in our experiment. Table~\ref{tab:hyperparameter} details the specific generation parameters for the various MLLMs we evaluate.

\begin{figure*}[ht]
\centering
\includegraphics[width=0.88\textwidth]{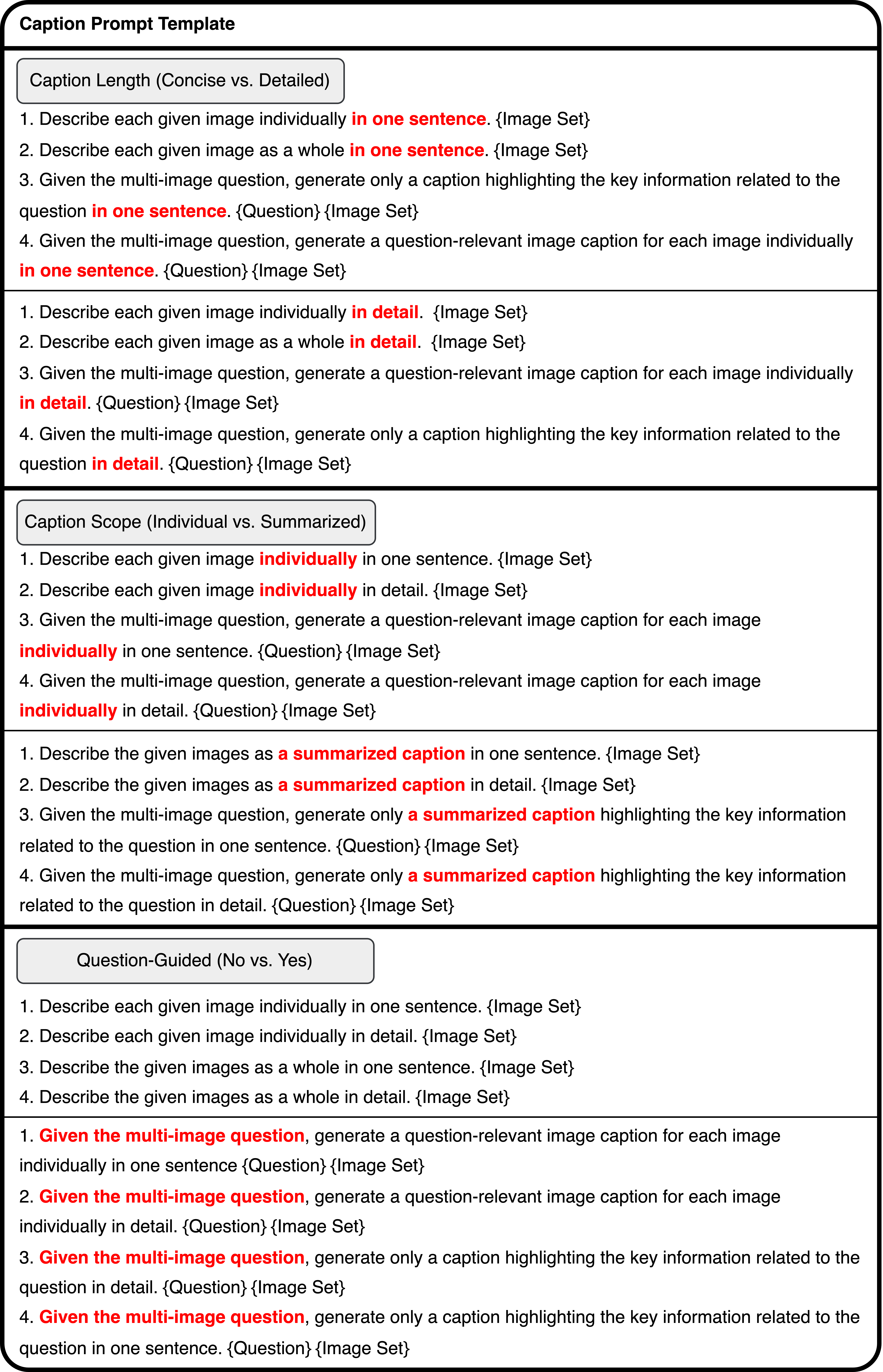}
\vspace{-2mm}
\captionsetup{font=small, width=0.9\linewidth}
\caption{Actual prompts with different captioning settings.}
\label{fig:different_caption_prompt}
\vspace{-4mm}
\end{figure*}

\begin{figure*}[ht]
\centering
\includegraphics[width=0.73\textwidth]{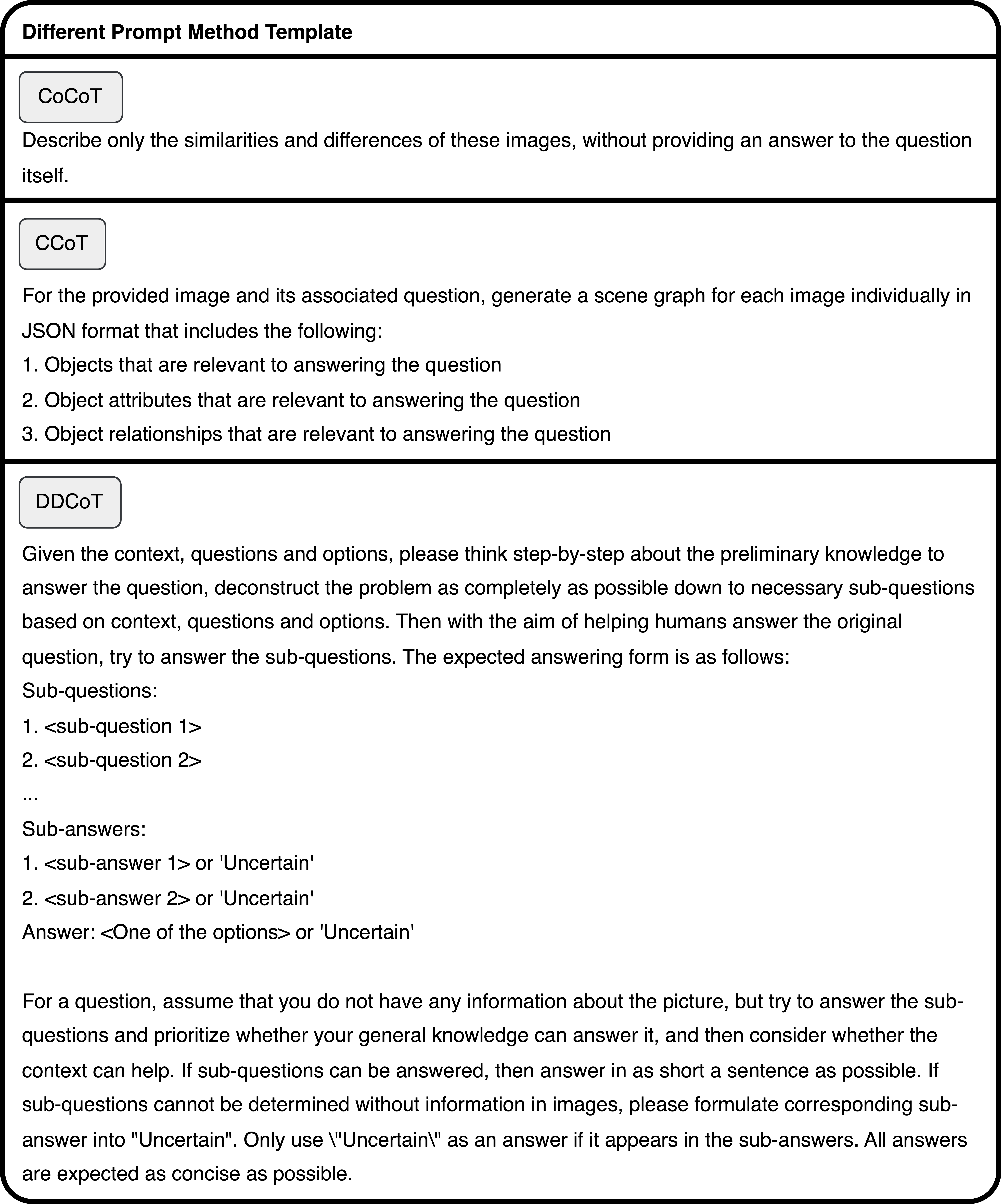}
\vspace{-2mm}
\captionsetup{font=small, width=0.9\linewidth}
\caption{Different actual prompts of existing prompting methods adapted to multi-image scenarios.}
\label{fig:different_method_prompt}
\vspace{-4mm}
\end{figure*}

\begin{figure*}[ht]
\centering
\includegraphics[width=0.73\textwidth]{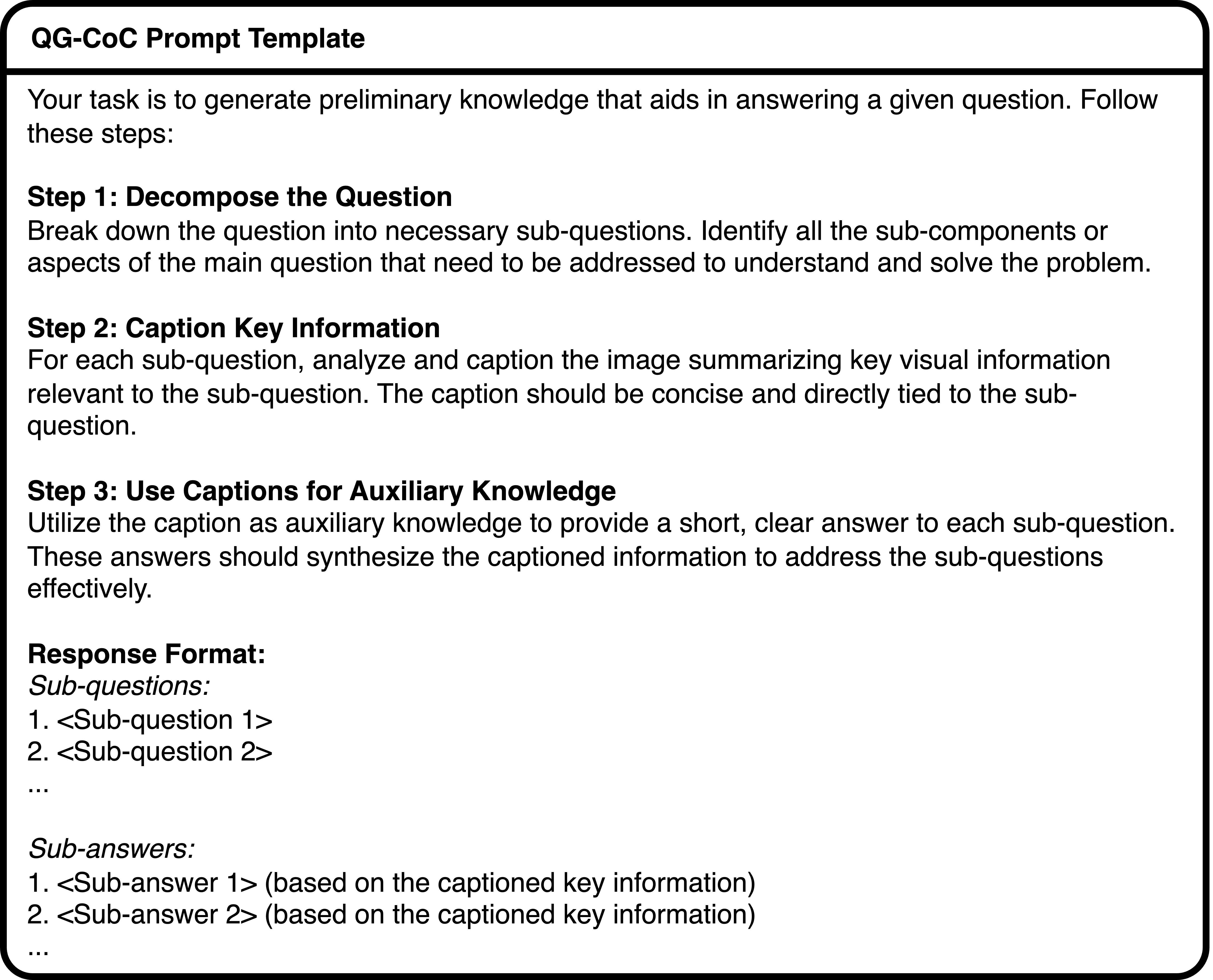}
\vspace{-2mm}
\captionsetup{font=small, width=0.9\linewidth}
\caption{An actual prompt of QG-CoC.}
\label{fig:qgcoc_prompt}
\vspace{-2mm}
\end{figure*}

\section{Detail Studies of Different Captioning Strategies under Multi-Image}
\label{sec:different_captioning_strategy}
\subsection{Full Model Prompt}
In Figure~\ref{fig:different_caption_prompt}, we show the full model prompt of different captioning settings. 

% \subsection{More Qualitative Examples}
% In Figure~\ref{fig:examples_captioning_settings}, we show more examples for each multi-image task using different captioning strategies in Gemini-1.5-Flash.

% \begin{figure*}[h]
%  \centering
%   \begin{subfigure}{0.477\textwidth}
%     \includegraphics[width=\linewidth]{Figure/caption1.png}
%     \captionsetup{font=small, width=0.8\linewidth}
%     \caption{Task: Scene Understanding}
%     \label{fig:caption_1}
%   \end{subfigure}
%   \hfill
%   \begin{subfigure}{0.477\textwidth}
%     \includegraphics[width=\linewidth]{Figure/caption2.png}
%     \captionsetup{font=small, width=0.8\linewidth}
%     \caption{Task: Visual Retrieval}
%     \label{fig:caption_2}
%   \end{subfigure}
% \captionsetup{font=small, width=0.9\linewidth}
%  \caption{Examples of different tasks using different captioning strategies on Gemini-1.5-Flash. Text in red, green, and orange highlights our advantages. Text in blue is the correct answer.}
%  \label{fig:examples_captioning_settings}
% \end{figure*}

\section{Detail Studies of Adjusting Existing Prompting Methods to Multi-Image Scenarios}
\label{sec:different_prompt}
\subsection{Full Model Prompt}
In Figure~\ref{fig:different_method_prompt}, we show the full model prompt of different methods. 

% \subsection{More Qualitative Examples}
% In Figure~\ref{fig:examples_prompt_method}, we show more examples for each multi-image task using different prompting methods in Gemini-1.5-Flash.

% \begin{figure*}[h]
%  \centering
%   \begin{subfigure}{0.477\textwidth}
%     \includegraphics[width=\linewidth]{Figure/different_prompt1.png}
%     \captionsetup{font=small, width=0.8\linewidth}
%     \caption{Task: Difference Spotting}
%     \label{fig:prompt_1}
%   \end{subfigure}
%   % \hfill % This command adds a horizontal space between the subfigures
%   % \begin{subfigure}{0.477\textwidth}
%   %   \includegraphics[width=\linewidth]{Figure/different_prompt2.png}
%   %   \captionsetup{font=small, width=0.8\linewidth}
%   %   \caption{Task: Visual Retrieval}
%   %   \label{fig:prompt_2}
%   % \end{subfigure}
%   % \begin{subfigure}{0.477\textwidth}
%   %   \includegraphics[width=\linewidth]{Figure/different_prompt3.png}
%   %   \captionsetup{font=small, width=0.8\linewidth}
%   %   \caption{Task: Image Text Matching}
%   %   \label{fig:prompt_3}
%   % \end{subfigure}
%   \hfill
%   \begin{subfigure}{0.477\textwidth}
%     \includegraphics[width=\linewidth]{Figure/different_prompt4.png}
%     \captionsetup{font=small, width=0.8\linewidth}
%     \caption{Task: Action Understanding}
%     \label{fig:prompt_4}
%   \end{subfigure}
% \captionsetup{font=small, width=0.9\linewidth}
%  \caption{Examples of different tasks using different prompting methods on Gemini-1.5-Flash. Text in red highlights the disadvantages. Text in blue is the correct answer.}
%  \label{fig:examples_prompt_method}
% \end{figure*}

\section{Detail Studies of Question-Guided Chain-of-Captions}
\label{sec:qgcoc}
\subsection{Full Model Prompt}
In Figure~\ref{fig:qgcoc_prompt}, we show the full model prompt of QG-CoC.

\subsection{Full Quantitative Results Across Various Image Relationships}
We further show the overall performance of QG-CoC across various image relationships and compare it with different prompting methods and models. The results of MMIU and MUIR datasets are shown in Table~\ref{tab:mmiu_prompting_comparison_detailed} and Table~\ref{tab:muir_prompting_comparison_detailed}, and we also illustrate the task performance of different prompting methods under MUIR benchmark in Figure~\ref{fig:examples_muir_all_dim_plot}. The findings remain the same as MMIU, and our method outperforms other methods. Additionally, we observe that the performance of each task under open-source models generally has a larger difference compared to closed-source models across various datasets and prompting methods.

% \subsection{More Qualitative Examples}
% In Figure~\ref{fig:examples_qgcoc}, we show more examples for each multi-image task using QG-CoC in Gemini-1.5-Flash.

% \begin{figure*}[h]
%  \centering
%   \begin{subfigure}{0.477\textwidth}
%     \includegraphics[width=\linewidth]{Figure/case1.png}
%     \captionsetup{font=small, width=0.8\linewidth}
%     \caption{Task: Image Text Matching}
%     \label{fig:case_1}
%   \end{subfigure}
%   \hfill % This command adds a horizontal space between the subfigures
%   \begin{subfigure}{0.477\textwidth}
%     \includegraphics[width=\linewidth]{Figure/case2.png}
%     \captionsetup{font=small, width=0.8\linewidth}
%     \caption{Task: Ordering}
%     \label{fig:case_2}
%   \end{subfigure}
  % \begin{subfigure}{0.7\textwidth}
  %   \includegraphics[width=\linewidth]{Figure/case3.png}
  %   \captionsetup{font=small, width=0.8\linewidth}
  %   \caption{Task: Screen Understanding}
  %   \label{fig:case_3}
  % \end{subfigure}
  % \hfill
  % \begin{subfigure}{0.477\textwidth}
  %   \includegraphics[width=\linewidth]{Figure/case4.png}
  %   \captionsetup{font=small, width=0.8\linewidth}
  %   \caption{Task: Visual Grounding}
  %   \label{fig:case_4}
  % \end{subfigure}
% \captionsetup{font=small, width=0.9\linewidth}
%  \caption{Examples of different tasks using QG-CoC on Gemini-1.5-Flash.}
%  \label{fig:examples_qgcoc}
% \end{figure*}

\subsection{Qualitative Analysis of Error Cases}
\begin{figure*}[t]
\centering
\begin{subfigure}[t]{0.47\textwidth}
    \includegraphics[width=\linewidth]{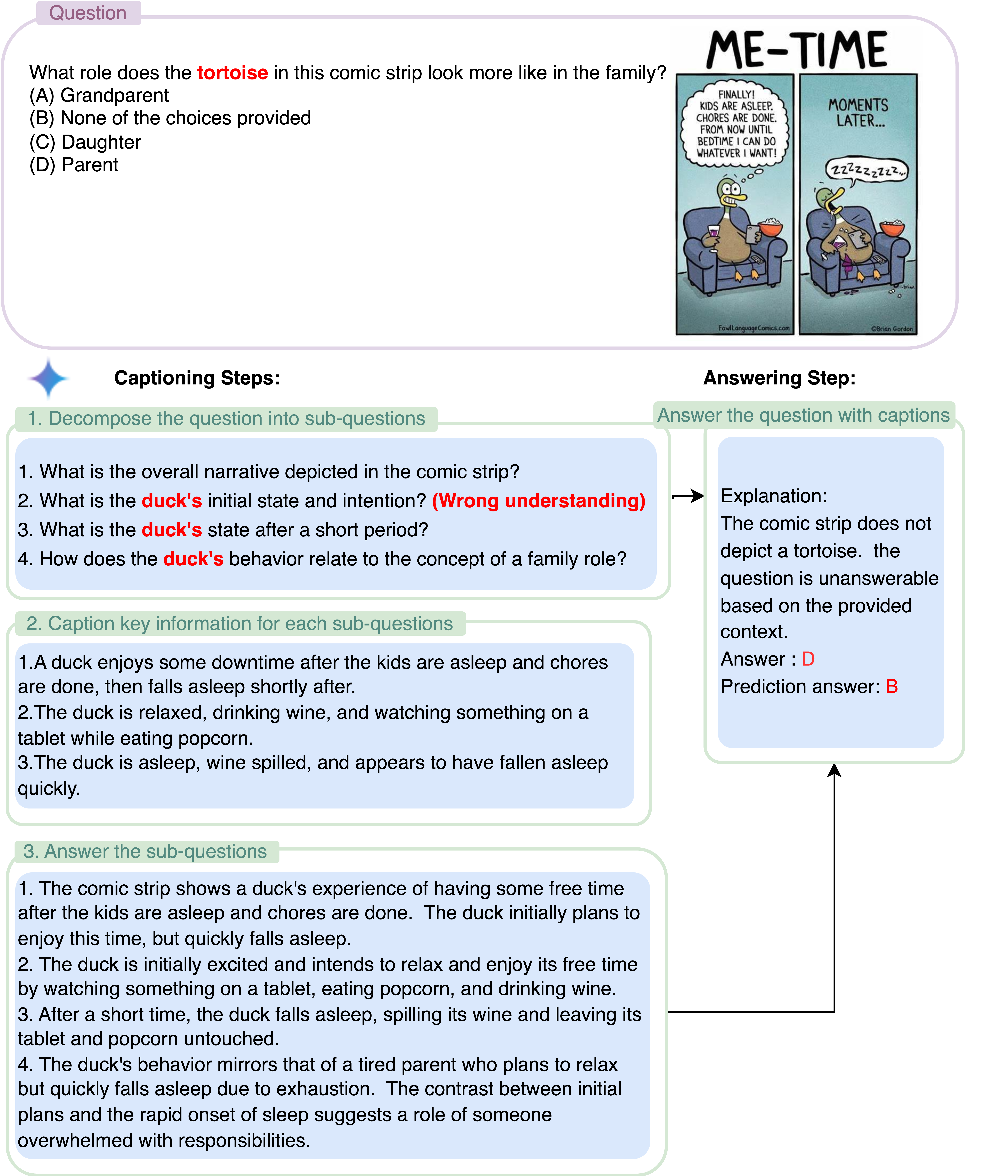}
    \caption{Error type 1 (Wrong Question Understanding) example of QG-CoC on Gemini-1.5-Flash.}
    \label{fig:error_case_1}
\end{subfigure}
\hfill
\begin{subfigure}[t]{0.47\textwidth}
    \includegraphics[width=\linewidth]{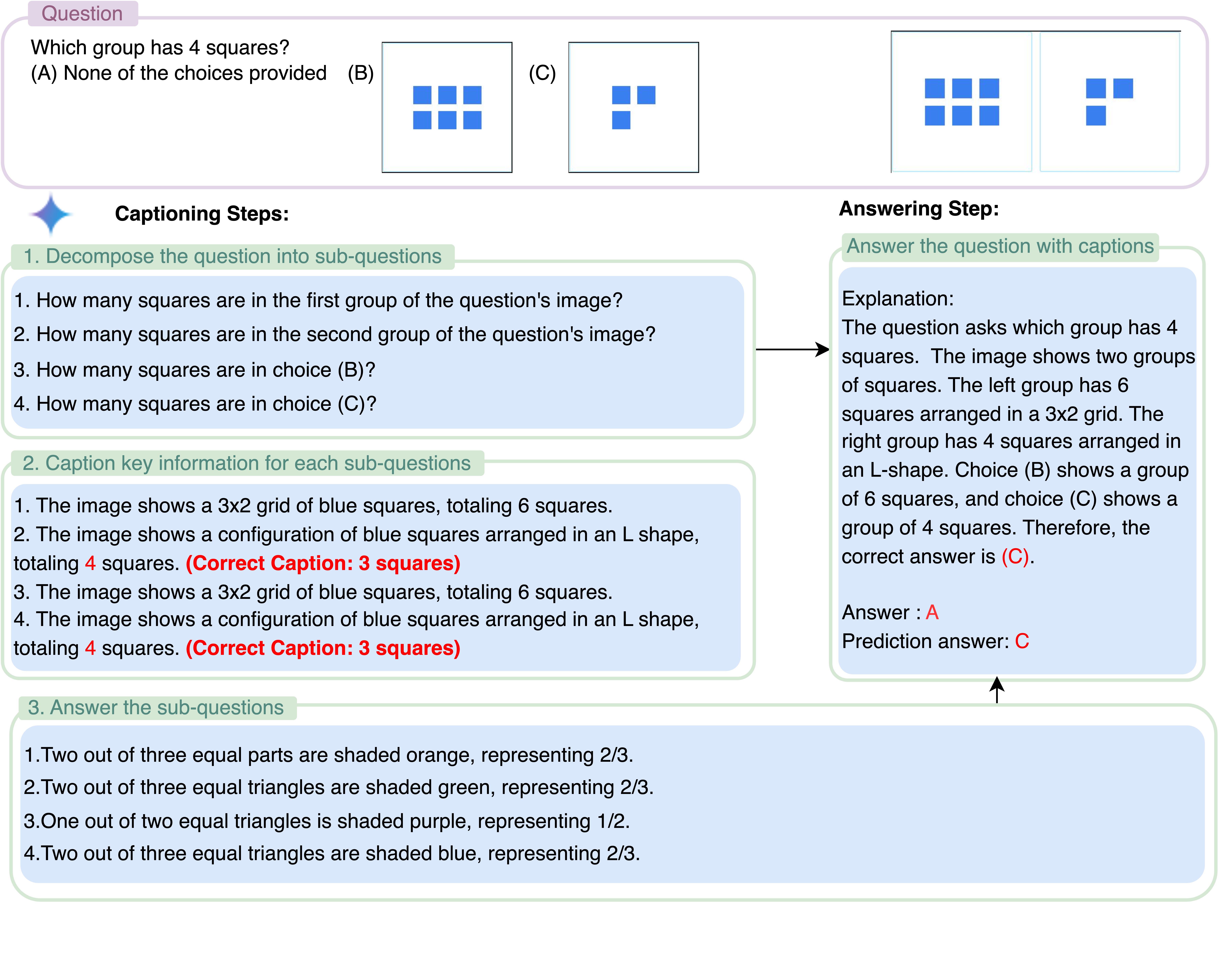}
    \caption{Error type 2 (Inaccurate Perception) example of QG-CoC on Gemini-1.5-Flash.}
    \label{fig:error_case_2}
\end{subfigure}
\begin{subfigure}[t]{0.85\textwidth}
    \includegraphics[width=\linewidth]{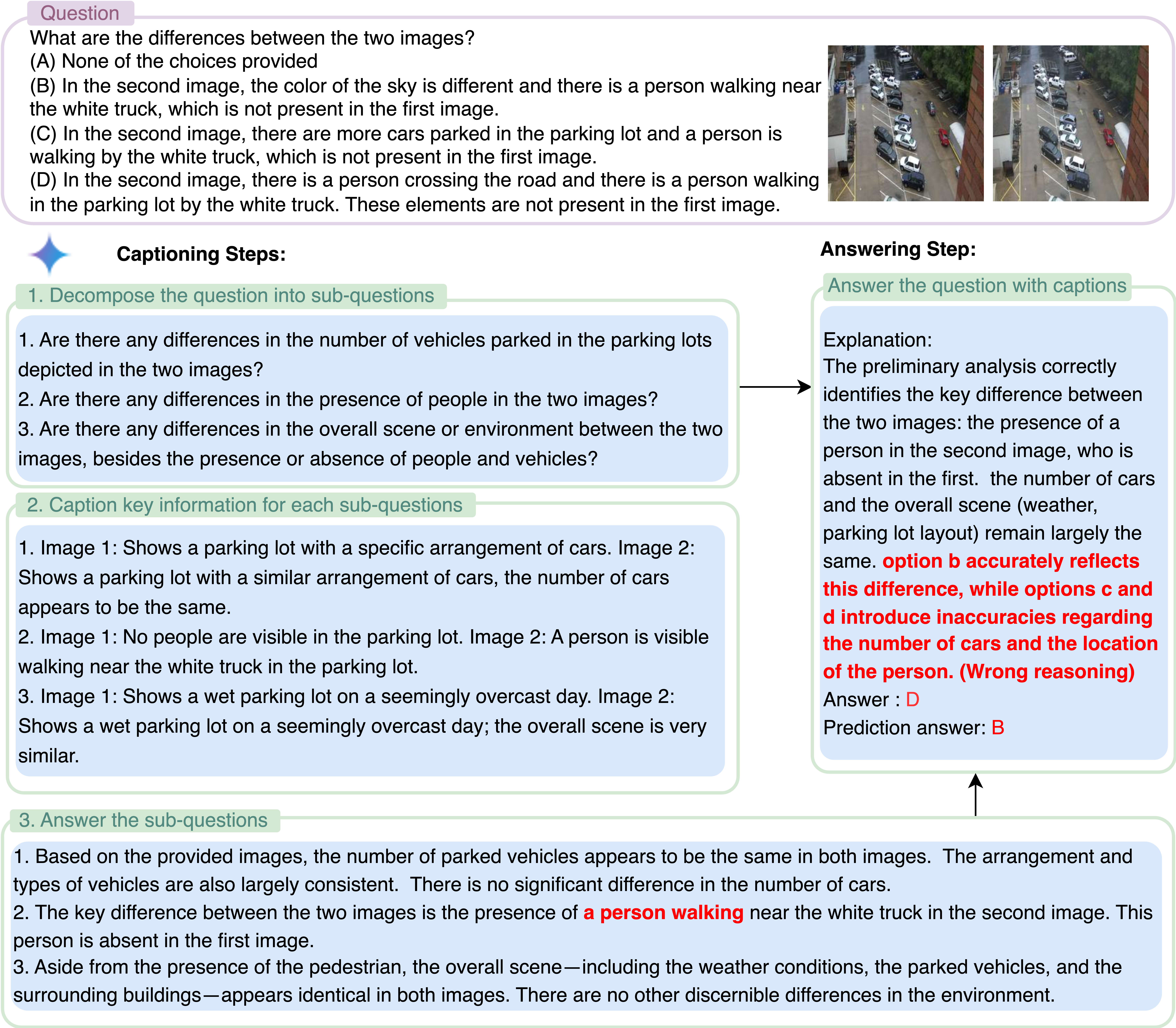}
    \caption{Error type 3 (Wrong Reasoning) example of QG-CoC on Gemini-1.5-Flash.}
    \label{fig:error_case_3}
\end{subfigure}
\captionsetup{font=small, width=0.9\linewidth}
\caption{Examples of three common error types made by QG-CoC on Gemini-1.5-Flash.}
\label{fig:error_cases}
\vspace{-2mm}
\end{figure*}

We present every type of error case that Gemini-1.5-Flash cannot answer correctly in Figure~\ref{fig:error_case_1}, \ref{fig:error_case_2},\ref{fig:error_case_3}. From E1, the model understands the wrong meaning of the question that "tortoise" is not "duck", and decomposes the question into wrong sub-questions (sub-goals). From E2, in step 2, the model incorrectly captions that "L shape has 4 squares", when the correct caption is "3 squares". From E3, since the generated sub-questions and captions are accurate, we can observe that the model correctly points out the difference between the two images, "a person walking". However, the model does incorrect reasoning in the final response.

% \begin{figure*}[h]
% \centering
% \includegraphics[width=0.4\textwidth]{Figure/error_case3.png}
% \vspace{-2mm}
% \captionsetup{font=small, width=0.9\linewidth}
% \caption{Error type 1 (Wrong Question Understanding) example of QG-CoC on Gemini-1.5-Flash.}
% \label{fig:error_case_1}
% \vspace{-2mm}
% \end{figure*}

% \begin{figure*}[ht]
% \centering
% \includegraphics[width=0.4\textwidth]{Figure/error_case1.png}
% \vspace{-2mm}
% \captionsetup{font=small, width=0.9\linewidth}
% \caption{Error type 2 (Inaccurate perception) example of QG-CoC on Gemini-1.5-Flash.}
% \label{fig:error_case_2}
% \vspace{-2mm}
% \end{figure*}

% \begin{figure*}[ht]
% \centering
% \includegraphics[width=0.4\textwidth]{Figure/error_case2.png}
% \vspace{-2mm}
% \captionsetup{font=small, width=0.9\linewidth}
% \caption{Error type 3 (Wrong reasoning) example of QG-CoC on Gemini-1.5-Flash.}
% \label{fig:error_case_3}
% \vspace{-2mm}
% \end{figure*}

\begin{table*}[ht]
\centering
\begin{adjustbox}{width=0.8\textwidth}
\begin{tabular}{c c |cccccccc}
\toprule
\textbf{Model} & \textbf{Method} & \textbf{Overall} & \textbf{Discrete} & \textbf{Continuous} & \textbf{Low-level} & \textbf{High-sub} & \textbf{High-obj} & \textbf{Two-D} & \textbf{Three-D} \\
\midrule

% --- LLaVA-One-Visit (llava-ov) Data ---
LLaVA-OV & w/o prompt & 44.6 & 37.6 & 47.9 & 66.8 & 51.8 & 42.9 & 37.1 & 27.8 \\
         & Caption    & 48.1 & 40.5 & 50.6 & 75.6 & 55.8 & 51.2 & 35.8 & 27.5 \\
         & QG-Caption & 49.4 & 40.1 & 53.4 & 78.4 & 56.3 & 53.8 & 37.6 & 26.5 \\
         & CCoT       & 50.5 & 41.4 & 50.2 & 76.9 & 57.5 & 59.1 & 39.6 & 28.5 \\
         & DDCoT      & 46.9 & 39.6 & 47.8 & 69.1 & 57.3 & 51.3 & 36.4 & 26.6 \\
         & CoCoT      & 46.4 & 39.6 & 48.0 & 72.3 & 53.5 & 48.2 & 36.5 & 26.8 \\
         & QG-CoC     & 50.9 & 39.4 & 52.3 & 71.9 & 60.0 & 61.0 & 37.8 & 34.1 \\
% \addlinespace
\midrule
% --- Mantis Data ---
Mantis & w/o prompt & 45.0 & 34.5 & 45.7 & 62.7 & 51.8 & 52.0 & 41.8 & 26.4 \\
       & Caption    & 46.7 & 35.4 & 45.7 & 69.5 & 52.0 & 52.7 & 40.7 & 28.6 \\
       & QG-Caption & 47.7 & 35.8 & 51.4 & 69.8 & 51.8 & 55.4 & 39.4 & 30.3 \\
       & CCoT       & 50.1 & 38.0 & 50.3 & 69.2 & 57.3 & 61.5 & 45.9 & 28.8 \\
       & DDCoT      & 44.9 & 37.9 & 48.5 & 57.3 & 50.8 & 52.2 & 42.5 & 25.4 \\
       & CoCoT      & 45.4 & 34.6 & 45.7 & 67.6 & 50.8 & 49.8 & 41.6 & 27.6 \\
       & QG-CoC     & 49.8 & 37.4 & 50.4 & 68.7 & 55.8 & 61.9 & 44.6 & 30.1 \\
% \addlinespace
\midrule
% --- GPT-4o Data ---
GPT-4o & w/o prompt & 63.3 & 60.6 & 60.7 & 94.8 & 60.0 & 67.3 & 53.3 & 46.4 \\
      & Caption    & 63.6 & 59.0 & 57.5 & 95.1 & 65.8 & 65.9 & 53.3 & 48.6 \\
      & QG-Caption & 65.1 & 58.1 & 61.4 & 93.1 & 66.0 & 67.7 & 55.8 & 53.5 \\
      & CCoT       & 60.9 & 53.4 & 60.0 & 91.7 & 60.8 & 63.7 & 53.4 & 43.0 \\
      & DDCoT      & 62.9 & 57.3 & 58.3 & 94.1 & 64.0 & 65.1 & 54.4 & 47.0 \\
      & CoCoT      & 64.5 & 60.3 & 60.9 & 95.4 & 65.8 & 65.0 & 56.3 & 48.0 \\
      & QG-CoC     & 65.8 & 59.3 & 61.4 & 93.3 & 66.0 & 68.5 & 56.2 & 55.9 \\
% \addlinespace
\midrule

% --- Gemini Data ---
Gemini-Flash & w/o prompt & 55.0 & 49.4 & 53.0 & 82.1 & 62.0 & 61.3 & 46.4 & 30.9 \\
       & Caption    & 53.7 & 51.4 & 52.1 & 83.1 & 60.3 & 63.3 & 47.2 & 18.4 \\
       & QG-Caption & 54.9 & 52.8 & 55.1 & 78.3 & 59.5 & 63.0 & 47.5 & 28.1 \\
       & CCoT       & 51.9 & 48.1 & 52.3 & 72.2 & 59.8 & 60.9 & 45.6 & 24.5 \\
       & DDCoT      & 51.5 & 47.8 & 51.6 & 80.4 & 58.8 & 61.4 & 42.4 & 18.4 \\
       & CoCoT      & 55.5 & 50.8 & 52.3 & 79.6 & 59.8 & 63.2 & 49.1 & 33.8 \\
       & QG-CoC     & 55.4 & 51.1 & 54.6 & 76.8 & 60.3 & 63.4 & 48.1 & 33.6 \\

\bottomrule
\end{tabular}
\end{adjustbox}
\captionsetup{font=small, width=0.9\linewidth}
\caption{MMIU performance across dimensions with different prompting methods and models.}
\label{tab:mmiu_prompting_comparison_detailed} % Changed label for clarity
\end{table*}
\begin{table*}[ht]
\centering
\begin{adjustbox}{width=\textwidth}
\begin{tabular}{c c |ccccccccccccc}
\toprule
\textbf{Model} & \textbf{Method} & \textbf{Overall} & \textbf{Geographic.} & \textbf{Diagram.} & \textbf{Matching.} & \textbf{Difference.} & \textbf{Retrieval.} & \textbf{Counting.} & \textbf{Attribute.} & \textbf{Scene.} & \textbf{Action.} & \textbf{Grounding.} & \textbf{Cartoon.} & \textbf{Ordering} \\
\midrule

% --- LLaVA-One-Visit (llava-ov) Data ---
LLaVA-OV & w/o prompt & 41.2 & 37.0 & 54.0 & 44.0 & 30.0 & 45.9 & 26.5 & 34.2 & 63.4 & 40.2 & 29.8 & 38.5 & 15.6 \\
         & Caption    & 42.0 & 46.0 & 56.0 & 44.0 & 32.4 & 38.4 & 34.2 & 28.6 & 66.7 & 42.1 & 32.1 & 37.2 & 20.3 \\
         & QG-Caption & 44.7 & 40.0 & 60.1 & 49.6 & 33.2 & 41.4 & 36.3 & 37.2 & 66.1 & 43.3 & 29.8 & 38.5 & 20.3 \\
         & CCoT       & 44.6 & 44.0 & 58.8 & 47.8 & 32.7 & 43.5 & 35.9 & 36.7 & 69.9 & 40.2 & 32.1 & 38.5 & 18.8 \\ % Data from image's CCoT row (5th data row)
         & DDCoT      & 53.4 & 41.0 & 69.6 & 61.0 & 46.2 & 54.5 & 34.2 & 56.1 & 74.2 & 42.1 & 32.1 & 41.0 & 21.9 \\ % Data from image's DDCoT row (4th data row)
         & CoCoT      & 44.2 & 42.0 & 56.8 & 46.3 & 34.4 & 50.3 & 31.6 & 35.7 & 67.2 & 42.1 & 31.0 & 35.9 & 17.2 \\
         & QG-CoC     & 53.3 & 42.0 & 70.1 & 60.1 & 38.8 & 54.1 & 41.9 & 56.6 & 76.9 & 43.9 & 29.8 & 42.3 & 20.3 \\
% \addlinespace
\midrule
% --- Mantis Data ---
Mantis & w/o prompt & 43.4 & 25.0 & 62.1 & 53.7 & 28.8 & 35.3 & 38.0 & 46.9 & 56.5 & 34.2 & 28.6 & 38.5 & 17.2 \\
       & Caption    & 43.9 & 29.0 & 61.3 & 53.0 & 32.7 & 31.9 & 39.3 & 33.7 & 62.9 & 44.5 & 28.6 & 43.6 & 17.2 \\
       & QG-Caption & 44.5 & 32.0 & 63.6 & 53.5 & 28.5 & 37.0 & 41.0 & 38.8 & 62.4 & 41.5 & 28.6 & 38.5 & 15.6 \\
       & CCoT       & 44.4 & 30.0 & 63.3 & 56.5 & 28.2 & 34.6 & 41.5 & 35.7 & 66.1 & 37.8 & 27.4 & 38.5 & 10.9 \\ % Data from image's CCoT row (5th data row)
       & DDCoT      & 47.9 & 35.0 & 59.8 & 57.8 & 35.9 & 42.1 & 39.3 & 52.0 & 71.0 & 38.4 & 34.5 & 41.0 & 15.6 \\ % Data from image's DDCoT row (4th data row)
       & CoCoT      & 42.6 & 26.0 & 59.6 & 52.6 & 33.8 & 31.5 & 39.3 & 35.2 & 55.9 & 38.4 & 29.8 & 38.5 & 17.2 \\
       & QG-CoC     & 48.9 & 37.0 & 64.3 & 59.1 & 34.5 & 41.4 & 44.0 & 48.0 & 70.4 & 39.0 & 32.1 & 46.2 & 15.6 \\
% \addlinespace
\midrule
% --- GPT-4o Data ---
GPT-4o & w/o prompt & 70.8 & 50.0 & 90.2 & 84.1 & 58.5 & 63.0 & 78.6 & 63.3 & 86.6 & 50.6 & 54.8 & 53.9 & 28.1 \\
      & Caption    & 71.8 & 62.0 & 91.0 & 85.6 & 65.3 & 59.9 & 79.1 & 56.1 & 83.3 & 54.9 & 53.6 & 52.6 & 34.4 \\
      & QG-Caption & 67.0 & 44.0 & 90.2 & 84.9 & 63.8 & 58.2 & 75.2 & 60.7 & 85.0 & 51.2 & 52.4 & 50.0 & 23.4 \\
      & CCoT       & 70.4 & 51.0 & 90.2 & 83.9 & 66.2 & 61.6 & 75.6 & 60.2 & 83.3 & 46.3 & 54.8 & 44.9 & 31.3 \\ % Data from image's CCoT row (5th data row)
      & DDCoT      & 73.1 & 50.0 & 89.7 & 85.8 & 66.5 & 64.4 & 79.9 & 61.7 & 87.6 & 57.3 & 56.0 & 56.4 & 40.6 \\  % Data from image's DDCoT row (4th data row)
      & CoCoT      & 74.0 & 57.0 & 90.5 & 87.3 & 70.6 & 70.9 & 76.5 & 59.2 & 88.2 & 50.0 & 54.8 & 57.7 & 37.5 \\
      & QG-CoC     & 74.9 & 61.0 & 91.0 & 87.9 & 68.5 & 68.5 & 79.1 & 62.2 & 87.0 & 57.9 & 57.1 & 56.4 & 43.8 \\
% \addlinespace
\midrule

% --- Gemini Data ---
Gemini-Flash & w/o prompt & 66.0 & 53.0 & 84.7 & 82.5 & 53.5 & 75.3 & 51.3 & 54.1 & 82.8 & 43.3 & 51.2& 46.2 & 18.8 \\
       & Caption    & 66.9 & 58.0 & 84.2 & 83.2 & 56.2 & 69.2 & 50.9 & 58.2 & 80.7 & 47.6 & 50.0 & 50.0 & 32.8 \\
       & QG-Caption & 66.0 & 47.0 & 83.4 & 83.4 & 55.0 & 64.4 & 52.1 & 61.2 & 83.3 & 53.1 & 48.8 & 42.3 & 25.0 \\
       & CCoT       & 66.3 & 54.0 & 85.7 & 82.3 & 52.4 & 69.9 & 50.0 & 60.7 & 81.2 & 49.4 & 47.6 & 43.6 & 34.4 \\ % Data from image's CCoT row (5th data row)
       & DDCoT      & 67.6 & 44.0 & 87.7 & 84.3 & 56.5 & 74.7 & 46.6 & 62.2 & 75.8 & 49.4 & 56.0 & 53.9 & 32.8 \\ % Data from image's DDCoT row (4th data row)
       & CoCoT      & 65.4 & 44.0 & 84.4 & 81.7 & 50.9 & 73.3 & 48.7 & 57.1 & 80.7 & 47.0 & 51.2 & 52.6 & 25.0 \\
       & QG-CoC     & 68.2 & 46.0 & 88.7 & 84.3 & 57.4 & 76.0 & 50.4 & 59.2 & 79.0 & 50.6 & 52.4 & 51.3 & 28.1 \\

\bottomrule
\end{tabular}
\end{adjustbox}
\captionsetup{font=small, width=0.9\linewidth}
\caption{MUIR performance across tasks with different prompting methods and models.}
\label{tab:muir_prompting_comparison_detailed}
\end{table*}

\end{document}